\pdfoutput=1

\documentclass[11pt]{article}

\usepackage[final]{acl}

\usepackage{times}
\usepackage{latexsym}

\usepackage[T1]{fontenc}

\usepackage[utf8]{inputenc}

\usepackage{microtype}

\usepackage{inconsolata}

\usepackage{graphicx}
\usepackage{amsmath}
\usepackage{graphicx}
\usepackage{dsfont}
\usepackage{caption}
\usepackage{tabularx}
\usepackage{wrapfig}
\usepackage{multirow}
\usepackage{colortbl}  
\usepackage{listings}
\usepackage{collcell}
\usepackage{amssymb}
\usepackage{array}
\usepackage{enumitem}
\usepackage{subcaption}
\usepackage{booktabs}
\usepackage{makecell}

\DeclareMathOperator*{\argmax}{arg\,max}

\newcommand{\method}{\textsc{NeKo}}
\newcommand{\methodwithspace}{\textsc{NeKo }}
\title{\raisebox{-0.20\height}{\includegraphics*[width=0.85cm]{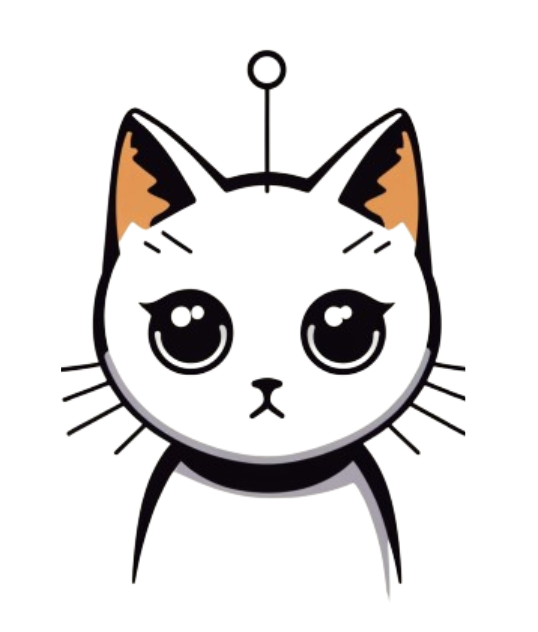}}\textsc{NeKo}: Cross-Modality Post-Recognition Error Correction with Tasks-Guided Mixture-of-Experts Language Model}

\author{\begin{tabular}{@{}c@{}c@{}}
Yen-Ting Lin$^{}$\thanks{Work done at NVIDIA research as an intern.}\quad Zhehuai Chen\quad Piotr Zelasko\quad Zhen Wan\quad Xuesong Yang\\ Zih-Ching Chen\quad Krishna C Puvvada\quad Szu-Wei Fu \quad Ke Hu \quad Jun Wei Chiu \\ \quad Jagadeesh Balam \quad Boris Ginsburg \quad Yu-Chiang Frank Wang \quad Chao-Han Huck Yang 
\end{tabular}\\NVIDIA
\\{\tt corresponding authors: ytl@ieee.org, hucky@nvidia.com}}

\begin{document}
\maketitle
\begin{abstract}
Construction of a general-purpose post-recognition error corrector poses a crucial question: how can we most effectively train a model on a large mixture of domain datasets? The answer would lie in learning dataset-specific features and digesting their knowledge in a single model. Previous methods achieve this by having separate correction language models, resulting in a significant increase in parameters. In this work, we present Mixture-of-Experts as a solution, highlighting that MoEs are much more than a scalability tool. We propose a Multi-Task Correction MoE, where we train the experts to become an ``expert'' of speech-to-text, language-to-text and vision-to-text datasets by learning to route each dataset's tokens to its mapped expert. 
Experiments on the Open ASR Leaderboard show that we explore a \textbf{new state-of-the-art} performance by achieving an average relative $5.0$\% WER reduction and substantial improvements in BLEU scores for speech and translation tasks. On zero-shot evaluation, NeKo outperforms GPT-3.5 and Claude-3.5 Sonnet with $15.5$\% to $27.6$\% relative WER reduction in the Hyporadise benchmark. NeKo performs competitively on grammar and post-OCR correction as a multi-task model. 
\end{abstract}

\section{Introduction}

\begin{figure}[t!]
\begin{center}
\includegraphics[width=1.1\columnwidth]{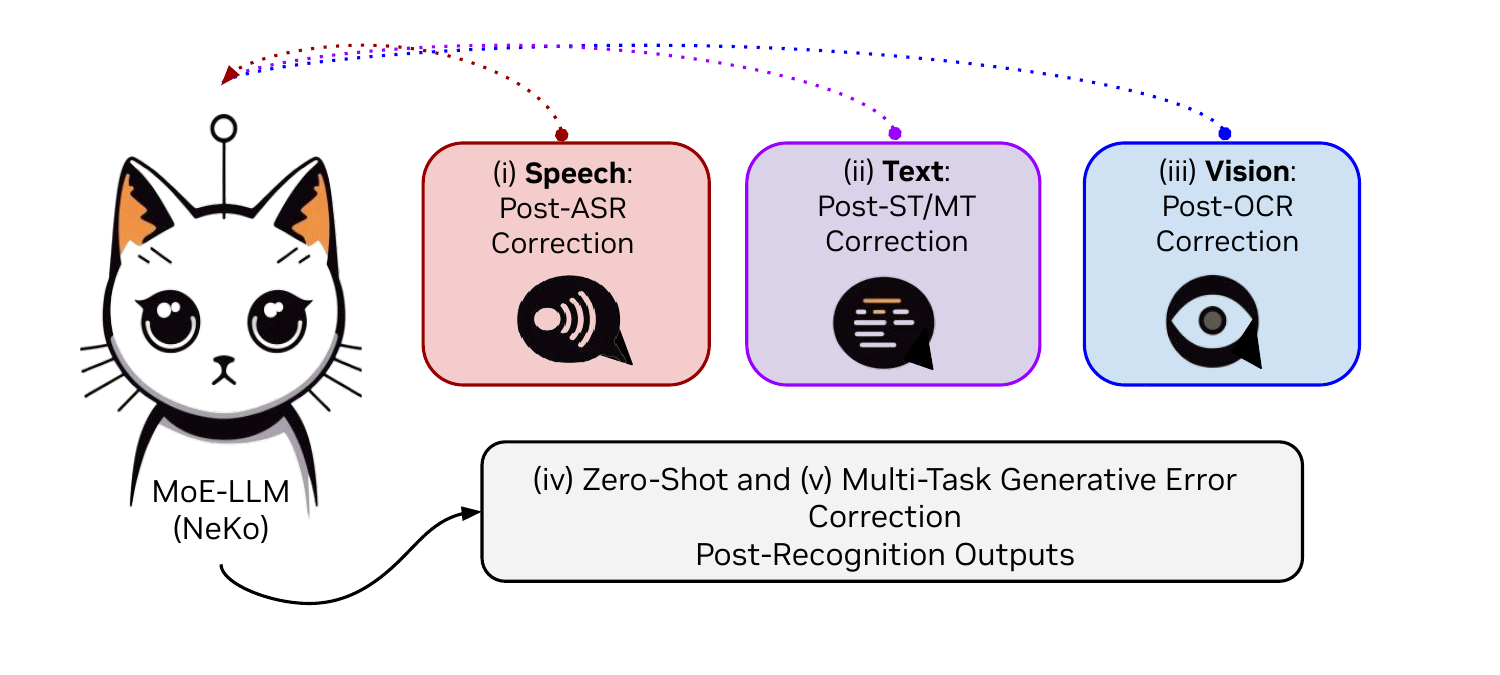}
\end{center}
\caption{Proposed \method, a new form multi-task model to boost post-recognition results over speech, text, and visual inputs. \methodwithspace could work for (i) post automatic speech recognition (ASR) correction, (ii) post speech translation (ST) and machine translation (MT) correction, and (iii) post optical character recognition (OCR) correction. NeKo discover new state-of-the-art results in (iv) zero-shot ASR correction and performs competitively as a general-purpose (v) multi-task corrector.} 
\label{fig:sytem}
\end{figure}

Human recognition capabilities span multiple modalities, including speech recognition, visual patterns, and extensions to semantic and textual interpretations. These faculties, however, are not infallible and often incorporate mis-recognition errors. Despite these imperfections, humans efficiently communicate using speech, language, or facial expressions.

For instance, two non-native speakers~\citep{lev2015comprehending, valaki2004cortical} can often achieve mutual understanding through this imperfect recognition and subsequent interpretative processes, even when the conversation is marred by lexical inaccuracies and subdued accents. In other words, humans (as intelligent agents) exhibit a robust capacity for generative understanding~\cite{jiang2020believe, cheng2021problematic} that extends beyond initial recognition results. In neuroscience~\cite{zatorre2008neural}, the inferior temporal gyrus and the temporal lobe are not confined to rudimentary perception but are also integral to the post-recognition processes that facilitate semantic understanding of language~\citep{levinson2010time}, speech~\citep{marshall2015deaf}, and visual patterns~\citep{vink2020towards}. This form of ``post-recognition correction,'' exemplified by the application of language modeling (LM) to initial recognition outputs, has been introduced to the field for both acoustic (automatic speech recognition, ASR) and visual (optical character recognition, OCR) modalities.


With the LMs scaling up to LLMs~\citep{Brown2020LanguageMA}, recent efforts~\citep{chan2023ic3,yang2023generative,chen2023hyporadise,hu2024large} have focused on exploring a ``generative modeling'' for post-recognition correction. This generative error correction (GER) approach uses LLMs to conduct final recognition from given first-pass text-based predictions from recognition models, including ASR, image captioning (IC), and machine translation (MT). This cascaded two-agents text-to-text GER model has outperformed larger single multi-modal and multi-task models in these tasks. Meanwhile, these GER solutions heavily depends on domain-specific fine-tuning processes~\citep{chen2024s} that utilize parameter-efficient components, which often suffers a performance \textit{degradation} from a lack of generalizability across different datasets, domains, and tasks. 




To characterize ``model generalization,'' mixture-of-experts (MoE) \citep{DBLP:journals/corr/abs-2401-04088} has emerged as a promising approach for multi-task learning, consisting of of a set of \textit{expert networks} and a \textit{gating network} that learns to route the input to the most appropriate expert~\citep{DBLP:journals/corr/abs-2403-07816}. This enables MoE models to learn more specialized and fine-grained representations compared to monolithic models.
However, most MoE models are designed for general-purpose language modeling\citep{DBLP:journals/corr/abs-2401-06066}, with experts not explicitly assigned to specific tasks, but rather learn to specialize in different aspects of the input space through data-driven training. Effectively leverage MoE for multi-task error correction, where the experts need to capture task-specific features while allowing knowledge sharing, remains an open question.

In this work, we propose \method, a ``ge\textsc{Ne}rative multi-tas\textsc{K} error c\textsc{o}rrection'' approach that leverages a pre-trained MoE model to drive diverse tasks and cross-domain knowledge, as shown in Figure~\ref{fig:sytem}. The key idea is to continuously pre-train MoE model on a mixture of error correction datasets, with each expert specializing in a specific domain. This task-guided MoE fine-tuning approach enables the experts to capture task-specific features while allowing knowledge sharing through the router. We further pursue this direction by modeling MoE on error correction and highlight the effectiveness and robustness of MoEs in learning from a mixture of correction datasets.



\methodwithspace captures the nuances of each task, benefiting from shared knowledge across experts. Evaluated on tasks such as ASR, ST, OCR, and unseen textual error correction (TEC), \methodwithspace consistently outperforms baseline models, including Claude-3.5 Sonnet and GPT-3.5. It achieves state-of-the-art WER reduction on the Hyporadise benchmark and large-scale Open ASR Leaderboard~\citep{open-asr-leaderboard}. \methodwithspace also significant improves in OCR error correction. Further analysis confirms its robust multi-task capabilities. In summary, the main contributions of this work include:
\begin{enumerate}[left=1pt]
    \item We introduce \method, a multi-task error correction LLM that leverages task-guided mixture-of-experts for diverse post-recognition correction tasks. To the best of our knowledge, this is the first work that explores the use of MoE for multi-task error correction.
    
    \item \method\ has been studied under a new form of cross-modalities post-recognition correction evaluation, 
    serving as strong open-source ASR, ST, OCR, and TEC baselines. Our results show that \method\ discovers new state-of-the-art performance in ASR as a multi-task correction model.

    \item We discovered emergent abilities for cross-task correction from \methodwithspace as a first-of-its-kind multi-task correction approach toward a general-purpose post-recognition LM designs.
    \item The \methodwithspace models, newly created source datasets, and training processes are scheduled to open source under the CC BY-SA 4.0 license to support reproducibility in future research.

\end{enumerate}

\section{Method}

\begin{figure}[t]
    \centering
    \includegraphics[width=0.5\textwidth]{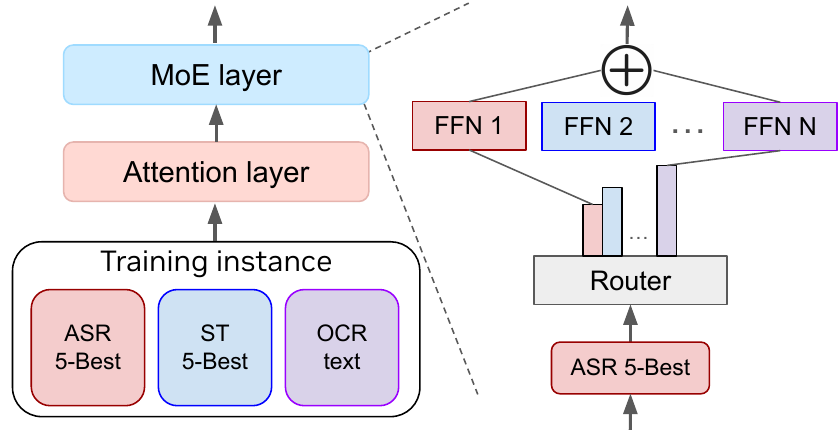}
    \caption{
        The architecture of our proposed model, \method, which integrates MoE layers within a Transformer architecture. 
        During inference, we do not assume knowledge of the specific task an input belongs to and each token is routed to the top-$2$ experts solely based on their router probabilities. 
    }
    \label{fig:moe_attention_architecture}
\end{figure}

\subsection{Mixture-of-Experts (MoE)}
Our method, \method, is based on a Transformer architecture \citep{vaswani2017attention} with modifications similar to those described in \citet{jiang2023mistral}. The key difference is that we replace the feedforward blocks with Mixture-of-Expert (MoE) layers.
In a MoE layer, each input token is assigned to a subset of experts by a gating network (router). The output of the MoE layer is the weighted sum of the outputs of the selected experts, where the weights are determined by the gating network. Formally, given $n$ expert networks $\{E_0, E_1, ..., E_{n-1}\}$, the output of the MoE layer for an input token $x$ is:

\begin{equation}
    y = \sum_{i=0}^{n-1} G(x)_i \cdot E_i(x),
\end{equation}

where $G(x)_i$ is the weight assigned to the $i$-th expert by the gating network, and $E_i(x)$ is the output of the $i$-th expert network for input $x$. The gating network $G(x)$ is implemented as a softmax over the top-$K$ logits of a linear layer:

\begin{equation}
    G(x) = \text{Softmax}(\text{TopK}(x \cdot W_g)),
\end{equation}

where $\text{TopK}(\ell)_i = \ell_i$ if $\ell_i$ is among the top-$K$ coordinates of logits $\ell \in \mathbb{R}^n$, and $\text{TopK}(\ell)_i = -\infty$ otherwise. The number of experts $K$ used per token is a hyperparameter that controls the computational cost. 


\subsection{Tasks-Guided Auxiliary Expert Assignment}
The key idea of \methodwithspace is to assign each expert to a specific task during training. Given a set of tasks $\mathcal{T} = \{T_1, T_2, ..., T_m\}$, we define a mapping function $f: \mathcal{T} \rightarrow \{1, 2, ..., n\}$ that assigns each task to a unique expert. During training, for an input token $x$ from task $T_i$, we deterministically route $x$ to the expert $f(T_i)$ in addition to the top-1 expert selected by the gating network. This ensures that each expert learns task-specific features while still allowing for knowledge sharing through the gating network.
Formally, the output of the MoE layer for an input token $x$ from task $T_i$ during training is:
\begin{equation}
    y = G(x)_{f(T_i)} \cdot E_{f(T_i)}(x) + G(x)_{\text{top1}} \cdot E_{\text{top1}}(x),
\end{equation}
where $\text{top1} = \argmax_{j \neq f(T_i)} G(x)_j$ is the index of the top-1 expert selected by the gating network, excluding the task-specific expert $f(T_i)$.

During inference, we do not assume knowledge of the specific task an input token belongs to. Instead, we route each token to the top-$K$ experts selected by the gating network based on their predicted probabilities. This approach allows the model to leverage the task-specific knowledge learned by the experts during training while still being able to generalize to new, potentially unseen tasks and domains during inference.

\subsection{Training Objective}
We train \methodwithspace on a mixture of error correction datasets $\mathcal{D} = \{D_1, D_2, ..., D_m\}$, where each dataset $D_i$ corresponds to a specific task $T_i$. The training objective is to minimize the negative log-likelihood of the target sequences:

\begin{equation}
    \mathcal{L} = -\sum_{i=1}^m \sum_{(x, y) \in D_i} \log p(y|x, T_i),
\end{equation}

where $x$ is the input sequence (e.g., ASR hypotheses, OCR output), $y$ is the target sequence (e.g., ground-truth transcription, corrected text), and $p(y|x, T_i)$ is the probability of the target sequence given the input sequence and the task prompt (Figure \ref{fig:task_prompt}.)
By jointly training on multiple error correction datasets with task-guided expert assignment, \methodwithspace learns to capture task-specific features while allowing for knowledge sharing across tasks through the shared gating network and other model components.

\section{Experiments}
\subsection{Training and Evaluation Datasets}
\paragraph{ASR} To assess the ability to handle diverse and noisy real-world speech, we use the Open ASR Leaderboard \citep{DBLP:journals/corr/abs-2210-13352,open-asr-leaderboard} for ASR evaluation, which comprises nine diverse datasets spanning various domains and speaking styles. These include LibriSpeech \citep{panayotov15_libripseech}, Common Voice 9 \citep{DBLP:conf/lrec/ArdilaBDKMHMSTW20}, VoxPopuli \citep{wang21_voxpopuli}, TED-LIUM \citep{hernandez18_tedlium}, GigaSpeech \citep{chen21_gigaspeech}, SPGISpeech \citep{oneill21_kensho}, Earnings-22 \citep{delrio22_earnings22}, and AMI \citep{carletta07_ami, renals07_ami}, as one most representative benchmark due to its scale and data diversity.
We include the training set of above 8 datasets for NeKo training.
We use the word error rate as the evaluation metric for ASR.

\paragraph{ST and MT} For the translation error correction task, we use the subset of the HypoTranslate dataset \citep{DBLP:journals/corr/abs-2402-06894} for training and evaluation. This dataset includes translation  from FLEURS \citep{DBLP:conf/slt/ConneauMKZADRRB22}, CoVoST-2 \citep{DBLP:journals/corr/abs-2007-10310}, and MuST-C \citep{di-gangi-etal-2019-must}, covering a range of languages such as Spanish, French, Italian, Japanese, Portuguese, Chinese, and Persian. 

\paragraph{OCR} For the optical character recognition (OCR) error correction task, we use the en-us portion of the OCR dataset \citep{PleIAs_Post_OCR_Correction}, which contains newspaper texts from Chronicling America. 

\paragraph{TEC} For the textual error correction (TEC) task, we use a subset of the CoEdIT dataset \citep{raheja2023coedit} from Grammarly, which contains   82K task-specific instructions for text editing. 


\subsection{Task-Specific Recognition Systems and Baselines}
\label{assec:baselines}

\paragraph{ASR} We compare against state-of-the-art ASR models, \texttt{Whisper-V2-Large}~\citep{radford22_whipser}, \texttt{Canary}~\citep{canary} without applying GEC method. End-to-end ASR-LLM,   
\texttt{SALM}~\citep{chen2024salm}, Qwen2-audio, and Gemini-2-Flash have been also compared.
For all \textit{Cascaded ASR+GEC Methods}, the task-specific system is the Canary model.
This model transcribes the speech data and generate 5-best hypotheses for each utterance using temperature-based sampling \citep{DBLP:journals/cogsci/AckleyHS85} with $p=0.3$. This allows us to capture a diverse set of potential transcriptions for each utterance, which can be fed into our error correction model.

\paragraph{ST and MT} For the speech and machine translation tasks, we compare against state-of-the-art models \texttt{SeamlessM4T} \citep{barrault2023seamlessm4t}, \texttt{GenTranslate}\citep{hu2024gentranslate}, and cascaded approaches combining ASR and machine translation models (\textit{e.g.}, \texttt{Whisper + NLLB}~\citep{DBLP:journals/corr/abs-2207-04672}). These baselines cover both end-to-end speech translation models and pipeline approaches. We use SeamlessM4T-Large V2 as the task-specific system to decode $N$-best hypotheses from input speech by beam search algorithm. We did this in two steps by first transcribing the speech and then translating the text, following~\citep{hu2024gentranslate}. 
LLMs then take the N-best hypotheses to produce a final speech translation result.
To investigate the generalization of our model, we also evaluate it in an alternative scenario: a direct speech translation model, Canary, is used as the task-specific system to produce hypotheses. 

\paragraph{OCR and TEC} We compare our proposed method against two baselines: (1) the input text without any correction (denoted as \texttt{Baseline}) and (2) a Mixtral 8x7B model fine-tuned only on the respective dataset for each task (denoted as \texttt{Mistral 8x7B Direct Finetune}). This allows us to assess the effectiveness of our task-guided expert assignment approach in handling OCR and TEC errors, as its ability to leverage knowledge from multiple tasks to improve performance on individual tasks compared to direct fine-tuning on a single dataset.

\begin{table*}[ht!]
\caption{Cross-domain ASR correction results in zero-shot and few-shot settings on the Hyporadise benchmark \citep{chen2023hyporadise}. We compare \methodwithspace against GPT-4 Turbo and Claude-3.5 Sonnet in 0- and 5-shot settings. The baseline represents the WER of task-specific model \texttt{Whisper-Large}. The oracle results used in~\citet{chen2023hyporadise} (N-best and Compositional) provide an upper bound for the correction performance.}
\centering
\resizebox{2\columnwidth}{!}{
\begin{tabular}{c|c|c|cc|cc|ccc|cc }
\toprule[1.5pt]
Domain &\multirow{2}{*}{Test Set} &\multirow{2}{*}{Baseline} & \multicolumn{2}{c|}{GPT-3.5 Turbo} & \multicolumn{2}{c|}{Claude-3.5 Sonnet} & \multicolumn{3}{c|}{0-shot w/ \method} & \multicolumn{2}{c}{\emph{Oracle}} \\
Shift &&& 0-shot & 5-shot & 0-shot & 5-shot & \method-FFT & \method-BTX & \method-MoE & N-best & Comp. \\ \midrule
\multirow{3}{*}{\begin{tabular}[c]{@{}c@{}}Specific\\ Scenario\end{tabular} } &WSJ-\emph{dev93} &9.0 & $8.5_{\textcolor{teal}{-5.6\%}}$ & $7.7_{\textcolor{teal}{-14.4\%}}$ & $8.2_{\textcolor{teal}{-8.9\%}}$ & $7.4_{\textcolor{teal}{-17.8\%}}$ & $8.6_{\textcolor{teal}{-4.4\%}}$ & $7.5_{\textcolor{teal}{-16.7\%}}$ & $\textbf{6.8}_{\textcolor{teal}{-24.4\%}}$ & 6.5 & 5.3 \\
&WSJ-\emph{eval92} &7.6 & $7.3_{\textcolor{teal}{-3.9\%}}$ & $6.6_{\textcolor{teal}{-13.2\%}}$ & $7.0_{\textcolor{teal}{-7.9\%}}$ & $6.3_{\textcolor{teal}{-17.1\%}}$ & $7.4_{\textcolor{teal}{-2.6\%}}$ & $6.4_{\textcolor{teal}{-15.8\%}}$ & $\textbf{5.8}_{\textcolor{teal}{-23.7\%}}$ & 5.5 & 4.7 \\
&ATIS & 5.8 & $5.5_{\textcolor{teal}{-5.2\%}}$ & $5.0_{\textcolor{teal}{-13.8\%}}$ & $5.2_{\textcolor{teal}{-10.3\%}}$ & $4.7_{\textcolor{teal}{-19.0\%}}$ & $5.6_{\textcolor{teal}{-3.4\%}}$ & $4.8_{\textcolor{teal}{-17.2\%}}$ & $\textbf{4.2}_{\textcolor{teal}{-27.6\%}}$ & 3.5 & 2.4 \\\midrule
\multirow{4}{*}{\begin{tabular}[c]{@{}c@{}}Common\\ Noise\end{tabular} }&CHiME4-\emph{bus} & 18.8 & $17.6_{\textcolor{teal}{-6.4\%}}$& $16.2_{\textcolor{teal}{-13.8\%}}$ & $17.1_{\textcolor{teal}{-9.0\%}}$ & $15.7_{\textcolor{teal}{-16.5\%}}$ & $17.7_{\textcolor{teal}{-5.9\%}}$ & $15.9_{\textcolor{teal}{-15.4\%}}$ & $\textbf{14.5}_{\textcolor{teal}{-22.9\%}}$ & 16.8 & 10.7 \\
&CHiME4-\emph{caf} & 16.1 & $14.7_{\textcolor{teal}{-8.7\%}}$ &$13.7_{\textcolor{teal}{-14.9\%}}$ & $14.2_{\textcolor{teal}{-11.8\%}}$ & $13.2_{\textcolor{teal}{-18.0\%}}$ & $14.8_{\textcolor{teal}{-8.1\%}}$ & $13.4_{\textcolor{teal}{-16.8\%}}$ & $\textbf{12.2}_{\textcolor{teal}{-24.2\%}}$ & 13.3 & 9.1 \\
&CHiME4-\emph{ped} & 11.5 &$10.9_{\textcolor{teal}{-5.2\%}}$ &$9.7_{\textcolor{teal}{-15.7\%}}$& $10.5_{\textcolor{teal}{-8.7\%}}$ & $9.3_{\textcolor{teal}{-19.1\%}}$ & $11.0_{\textcolor{teal}{-4.3\%}}$ & $9.5_{\textcolor{teal}{-17.4\%}}$ & $\textbf{8.6}_{\textcolor{teal}{-25.2\%}}$ & 8.5 & 5.5 \\
&CHiME4-\emph{str} & 11.4 & $10.9_{\textcolor{teal}{-4.4\%}}$ & $9.7_{\textcolor{teal}{-14.9\%}}$ & $10.5_{\textcolor{teal}{-7.9\%}}$ & $9.3_{\textcolor{teal}{-18.4\%}}$ & $11.0_{\textcolor{teal}{-3.5\%}}$ & $9.4_{\textcolor{teal}{-17.5\%}}$ & $\textbf{8.5}_{\textcolor{teal}{-25.4\%}}$ & 9.0 & 6.0 \\ \midrule
\multirow{4}{*}{\begin{tabular}[c]{@{}c@{}}Speaker\\ Accent\end{tabular} }&MCV-\emph{af} & 25.3 &$24.9_{\textcolor{teal}{-1.6\%}}$ & $23.6_{\textcolor{teal}{-6.7\%}}$ & $24.4_{\textcolor{teal}{-3.6\%}}$ & $23.0_{\textcolor{teal}{-9.1\%}}$ & $25.0_{\textcolor{teal}{-1.2\%}}$ & $23.3_{\textcolor{teal}{-7.9\%}}$ & $\textbf{21.0}_{\textcolor{teal}{-17.0\%}}$ & 23.6 & 21.7 \\
&MCV-\emph{au} & 25.8 & $25.1_{\textcolor{teal}{-2.7\%}}$ & $24.0_{\textcolor{teal}{-7.0\%}}$& $24.6_{\textcolor{teal}{-4.7\%}}$ & $23.4_{\textcolor{teal}{-9.3\%}}$ & $25.2_{\textcolor{teal}{-2.3\%}}$ & $23.7_{\textcolor{teal}{-8.1\%}}$ & $\textbf{21.4}_{\textcolor{teal}{-17.1\%}}$ & 24.9 & 21.8 \\
&MCV-\emph{in} & 28.6 & $27.6_{\textcolor{teal}{-3.5\%}}$ & $25.0_{\textcolor{teal}{-12.6\%}}$ & $27.0_{\textcolor{teal}{-5.6\%}}$ & $24.3_{\textcolor{teal}{-15.0\%}}$ & $27.8_{\textcolor{teal}{-2.8\%}}$ & $24.6_{\textcolor{teal}{-14.0\%}}$ & $\textbf{22.2}_{\textcolor{teal}{-22.4\%}}$ & 27.1 & 22.6 \\
&MCV-\emph{sg} & 26.4 & $26.5_{\textcolor{gray}{+0.4\%}}$ & $25.1_{\textcolor{teal}{-4.9\%}}$& $25.9_{\textcolor{teal}{-1.9\%}}$ & $24.5_{\textcolor{teal}{-7.2\%}}$ & $26.6_{\textcolor{gray}{+0.8\%}}$ & $24.7_{\textcolor{teal}{-6.4\%}}$ & $\textbf{22.3}_{\textcolor{teal}{-15.5\%}}$ & 25.5 & 22.2 \\
\bottomrule[1.5pt]
\end{tabular}}
\label{ood-result}
\end{table*}

\begin{table*}[ht!]
\centering
\caption{ASR correction results on the Open ASR Leaderboard. We report the Word Error Rate (WER) for each dataset and the average across all 9 datasets. \methodwithspace establishes a new state-of-the-art performance on the leaderboard, outperforming both \textit{end-to-end ASR methods} and \textit{cascaded ASR+GEC approaches}. We report the actual tuning parameter in parentheses (.) and the sum of the frozen Whisper results in front.}
\resizebox{1\textwidth}{!}{%
\begin{tabular}{lc|c|cccccccccc}
\toprule
Model & Inference Para. & Avg. $\downarrow$ & AMI & Earnings22 & Gigaspeech & LS Clean & LS Other & SPGI & Tedlium & Voxp. & MCV9 \\
\midrule
\multicolumn{12}{l}{\emph{\textbf{ASR or SpeechLMs: End-to-end Voice Understanding Models}}} \\
Distil-Whisper-V2-L~\citep{gandhi2023distil} & 0.75B & 8.31 & 14.65 & 12.12 & 10.31 & 2.95 & 6.39 & 3.28 & 4.30 & 8.22 & 12.60 \\
Whisper-V2-L~\citep{radford22_whipser} & 1.5B & 8.06 & 16.82 & 12.02 & 10.57 & 2.56 & 5.16 & 3.77 & 4.01 & 7.50 & 10.11 \\
Canary~\citep{canary} & 2B & 6.67 & 14.00 & 12.25 & 10.19 & \textbf{1.49} & \textbf{2.49} & \textbf{2.06} & 3.58 & \textbf{5.81} & 7.75 \\
Bestow Speech LM~\citep{chen2024bestow} & 1.8B & \textbf{6.50} & \textbf{12.58} & 12.86 & 10.06 & 1.64 & 3.07 & 2.11 & \textbf{3.41} & 5.84 & \textbf{6.97} \\
Qwen2-Audio~\citep{chu2024qwen2} & 8B & 7.43 & - & - & - & 1.6 & 3.6 & - & - & - & - \\
Gemini-2.0-Flash & - & 8.56 & - & - & - & - & - & - & - & - & - \\
\midrule
\multicolumn{12}{l}{\emph{\textbf{ASR+LLM: \textcolor{blue}{Frozen} Whisper-v2-L (1.5B) + Voice Correction LMs}}} \\
+~Gemma 2B~\citep{team2024gemma} FFT & 3.5B (2B) & 6.61 & 13.20 & 12.30 & 10.40 & 1.60 & 2.60 & 2.20 & 3.70 & 6.00 & 7.50 \\
+~Gemma 8x2B FFT & 3.5B (2B) & 6.51 & 13.10 & 12.20 & 10.30 & 1.50 & 2.50 & 2.10 & 3.60 & 5.90 & 7.40 \\
+~\methodwithspace (Ours) Gemma 8x2B & 3.5B (2B) & 6.41 & 13.00 & 12.10 & 10.20 & 1.40 & 2.40 & 2.00 & 3.50 & 5.80 & 7.30 \\
\rowcolor{yellow!20}+~\methodwithspace (Ours) Qwen1.5-MoE & 4.2B (2.7B) & \textbf{5.90} & 12.60 & \textbf{11.82} & \textbf{9.95} & \textbf{1.30} & \textbf{2.32} & \textbf{1.94} & \textbf{3.20} & 5.80 & 7.30
\\\midrule
+~Mistral 7B~\citep{jiang2023mistral} FFT & 8.5B (7B) & 6.40 & 13.07 & 11.87 & 10.09 & 1.48 & 2.46 & 2.04 & 3.55 & \textbf{5.75} & 7.29 \\
+~Mixtral 8x7B~\citep{Jiang2024MixtralOE} FFT & 8.5B (7B) & 6.51 & 12.91 & 12.19 & 10.34 & 1.54 & 2.55 & 2.12 & 3.64 & 5.89 & 7.43 \\
+~Mixtral 8x7B Lora & 8.5B (7B) & 6.60 & 12.96 & 12.24 & 10.38 & 1.55 & 2.56 & 2.13 & 3.66 & 5.92 & 7.47 \\
+~Mistral 8x7B BTM~\citep{DBLP:journals/corr/abs-2403-07816} & 8.5B (7B) & 6.43 & 13.13 & 11.93 & 10.14 & 1.49 & 2.47 & 2.05 & 3.57 & 5.78 & 7.33 \\

\rowcolor{yellow!20}+~\methodwithspace (Ours) Mixtral 8x7B & 8.5B (7B) & \textbf{6.34} & \textbf{12.55} & \textbf{11.82} & 10.02 & 1.49 & 2.47 & 2.05 & 3.52 & 5.76 & \textbf{7.25} \\
+~\methodwithspace (Ours) Mixtral 8x22B & 23.5B (22B) & 6.40 & 12.61 & 11.93 & 10.15 & 1.52 & 2.51 & 2.09 & 3.58 & 5.82 & 7.33 \\
\bottomrule
\end{tabular}
}
\label{tab:multi_dataset_asr}
\end{table*}

\subsection{Post-recognition LLMs Setup}

We implement \methodwithspace using the Transformer architecture \citep{vaswani2017attention} and fine-tune both dense and MoE models for comparison. For dense models, we fine-tune Gemma 2B \citep{team2024gemma} and Mistral 7B \citep{Jiang2024MixtralOE}. For MoE models, we fine-tune Gemma 8x2B\footnote{We made an up-cycled~\citep{DBLP:conf/iclr/KomatsuzakiPLRM23} Gemma 8x2B MoE setup extended from single Gemma-2B~\citep{team2024gemma}.} and Mixtral 8x7B without applying our task-guided expert assignment.
We explore the Branch-Train-Mix approach \citep{DBLP:journals/corr/abs-2403-07816}, which involves branching from the Mistral 7B model to an 8x7B MoE model as one competing setup. 
To investigate the scalability of our method, we design \methodwithspace to three different sizes of MoE models: Gemma 8x2B, Mixtral 8x7B, and Mixtral 8x22B. We further compared low-rank adaptation (\texttt{LoRA}\citep{hu2021lora}) with full fine-tuning (\texttt{FFT}) on 8x7B MoE setup.

For  MoE models, we use top-k routing as proposed in \citep{DBLP:conf/iclr/LepikhinLXCFHKS21} to balance the computational cost and model capacity.
We use a global batch size of 2 million tokens and apply sample packing \citep{DBLP:journals/jmlr/RaffelSRLNMZLL20} to maximize the GPU utilization.


\begin{table*}[ht!]
\centering
\caption{Speech translation results on FLEURS, CoVoST-2, and MuST-C \textbf{En${\rightarrow}$X} test sets in terms of BLEU score.We use \textbf{bold} to highlight surpassing SeamlessM4T baseline, and use \underline{underline} to highlight the state-of-the-art performance. The baseline methods are introduced in \S\ref{assec:baselines}, and all of their results are reproduced by ourselves.}

\resizebox{1.0\textwidth}{!}{
\begin{tabular}{lccccccc|cccc|cccc}
\toprule[1.2pt]
\multirow{2}{*}{En$\rightarrow$X} & \multicolumn{7}{c|}{FLEURS} & \multicolumn{4}{c|}{CoVoST-2} & \multicolumn{4}{c}{MuST-C} \\
& Es & Fr & It & Ja & Pt & Zh & Avg. & Fa & Ja & Zh & Avg. & Es & It & Zh & Avg. \\
\midrule[1.2pt]
\multicolumn{16}{l}{\emph{\textbf{End-to-end ST Methods}}} \\
SeamlessM4T-Large~\citep{barrault2023seamlessm4t} & 23.8 & 41.6 & 23.9 & 21.0 & 40.8 & 28.6 & 30.0 & 18.3 & 24.0 & 34.1 & 25.5 & \textbf{34.2} & \textbf{29.9} & 16.2 & 26.8 \\
GenTranslate~\citep{hu2024gentranslate} & \textbf{25.4} & \textbf{43.1} & \textbf{25.5} & \textbf{28.3} & \textbf{42.4} & \textbf{34.3} & \textbf{33.2} & \textbf{21.1} & \textbf{29.1} & \textbf{42.8} & \textbf{31.0} & 33.9 & 29.4 & \textbf{18.5} & \textbf{27.3} \\
 SeamlessM4T-Large-V2~\citep{barrault2023seamlessv2} & 23.8 & 42.6 & 24.5 & 21.7 & 43.0 & 29.5 & 30.9 & 16.9 & 23.5 & 34.6 & 25.0 & 32.1 & \textbf{27.5} & 15.6 & 25.1 \\
 GenTranslate-V2~\citep{hu2024gentranslate} & \textbf{25.5} & \textbf{44.0} & \textbf{26.3} & \textbf{28.9} & \underline{\textbf{44.5}} & \textbf{34.9} & \textbf{34.0} & \textbf{19.4} & \textbf{29.0} & \underline{\textbf{43.6}} & \textbf{30.7} & \textbf{32.2} & 27.3 & \textbf{18.1} & \textbf{25.9} \\
\midrule
\multicolumn{16}{l}{\emph{\textbf{Cascaded ASR+MT Methods}}} \\
Whisper + NLLB-3.3b~\citep{DBLP:journals/corr/abs-2207-04672} & 25.1 & 41.3 & 25.0 & 19.0 & 41.5 & 23.5 & 29.2 & 13.6 & 19.0 & 32.0 & 21.5 & 35.3 & 29.9 & 13.5 & 26.2 \\
SeamlessM4T-Large (ASR+MT)~\citep{barrault2023seamlessm4t} & 24.6 & 44.6 & 25.4 & 22.5 & 41.9 & 31.2 & 31.7 & 18.8 & 24.0 & 35.1 & 26.0 & 35.1 & 30.8 & 17.7 & 27.9 \\
 SeamlessM4T-V2 (ASR+MT)~\citep{barrault2023seamlessv2} & 24.7 & 44.1 & 25.1 & 20.6 & 43.6 & 30.6 & 31.5 & 17.4 & 23.8 & 35.4 & 25.5 & 33.0 & 27.8 & 14.5 & 25.1 \\
\midrule
\multicolumn{16}{l}{\emph{\textbf{Cascaded ASR+GEC Methods}}} \\
GenTranslate & 26.8 & 45.0 & 26.6 & 29.4 & 43.1 & 36.8 & 34.6 & 21.8 & 30.5 & 43.3 & 31.9 & 35.5 & 31.0 & 19.6 & 28.7 \\
 GenTranslate-V2 & 27.0 & 44.3 & 26.4 & 27.8 & 44.5 & 36.1 & 34.4 & 20.8 & 29.7 & 43.5 & 31.3 & 33.2 & 28.3 & 16.9 & 26.1 \\
\method-Gemma-2B-FT & 26.9 & 44.2 & 26.3 & 27.7 & 44.4 & 36.0 & 34.3 & 20.7 & 29.6 & 43.4 & 31.2 & 33.1 & 28.2 & 16.8 & 26.0 \\
\method-Gemma-8x2B--BTX & 27.2 & 44.5 & 26.7 & 28.0 & 44.7 & 36.3 & 34.6 & 21.0 & 29.9 & 43.8 & 31.6 & 33.4 & 28.5 & 17.1 & 26.3 \\
\rowcolor{yellow!20}\method-Gemma-8x2B-MoE & \textbf{28.5} & \textbf{46.2} & \textbf{28.0} & \textbf{30.1} & \textbf{46.3} & \textbf{38.7} & \textbf{36.3} & \textbf{23.4} & \textbf{32.6} & \textbf{46.5} & \textbf{34.2} & \textbf{37.2} & \textbf{32.8} & \textbf{21.5} & \textbf{30.5} \\
\bottomrule[1.2pt]
\end{tabular}}
\label{tab:st_results}
\end{table*}

\subsection{Post-recognition Correction Results}

\paragraph{ASR} We first evaluate the zero-shot ability of \methodwithspace on unseen domain compared to two general-purpose LLMs, including GPT-3.5 Turbo and Claude-3.5 Sonnet. With a task-specific recognition baseline of Whisper-V2-Large (third column) in Table~\ref{ood-result}, \method-MoE (\textit{i.e.}, Qwen1.5-MoE or Mixtral) shows the best zero-shot ability with a relative 22.3\% average WER reduction. GPT-3.5 Turbo and Claude-3.5 Sonnet have relative 4.3\% and 7.3\% of zero-shot improvements, where \methodwithspace consistently outperform their 5-shot ASR correction.

Table \ref{tab:multi_dataset_asr} shows the WER scores on individual datasets and average performance on the Open ASR Leaderboard. We observe that the proposed \methodwithspace improves  the task-specific baseline Canary, with an average 5.0\% WER reduction.
Individually, we observe a significant performance increase with \methodwithspace on more challenging datasets, like AMI (conversational speech) and VoxPopuli (accented speech) due to experts learning dataset-specific features. While, Earnings22 shows a slight performance drop possibly due to the reduced representation in the batch. 

Compared to other leading models on the leaderboard, \methodwithspace establishes a new state-of-the-art, outperforming speech-only foundational models like Whisper and Canary and end-to-end ASR-LLM like SALM~\citep{chen2024salm} across most datasets. On the AMI dataset, \methodwithspace achieves a WER of 12.58\%, significantly lower than Whisper's 16.82\%. On VoxPopuli, \methodwithspace obtains 5.84\% WER, a 1.66 point reduction from Whisper's 7.5\%. 
The strong performance of \methodwithspace demonstrates the effectiveness of our speech-adapted MoE approach in handling diverse speech datasets and learning robust representations.

\paragraph{ST and MT} Table \ref{tab:st_results} presents the speech translation results on the FLEURS, CoVoST-2, and MuST-C datasets. For these experiments, we use \texttt{SeamlessM4T-Large} as the task-specific model to generate the initial speech translation hypotheses. \methodwithspace is then applied to correct the outputs from \texttt{SeamlessM4T-Large}. 
Compared to the task-specific \texttt{SeamlessM4T-Large} model, \methodwithspace achieves significant improvements, with an average BLEU score increase of 5.4 points on the FLEURS dataset, 9.2 points on the CoVoST-2 dataset, and 5.4 points on the MuST-C dataset. These results demonstrate the effectiveness of \methodwithspace in correcting errors made by the first-pass speech translation model.
Moreover, \methodwithspace outperforms other correction baselines, including the state-of-the-art \texttt{GenTranslate} model. 

\vspace{0.5cm}

\begin{table}[ht]
    \centering
    \caption{Machine translation BLEU scores on the WMT'20 Japanese (Ja) and Chinese (Zh) test sets \citep{DBLP:conf/wmt/BarraultBBCFGGH20}. \methodwithspace is evaluated in a zero-shot setting, while other models are fine-tuned on the respective language pairs. Higher BLEU scores indicate better translation quality.}
    \resizebox{1\columnwidth}{!}{
    \begin{tabular}{lccc}
        \toprule
        En$\rightarrow$X & WMT'20 Ja $\uparrow$ & WMT'20 Zh $\uparrow$ & Avg. $\uparrow$ \\
        \midrule
        ALMA-13b 
         & ~~3.5 & 11.3 & ~~7.4 \\
        BigTranslate  & ~~7.3 & 29.0 & 18.2 \\
        NLLB-3.3b  & 11.6 & 26.9 & 19.3 \\
        SeamlessM4T-Large & 17.0 & 27.0 & 22.0 \\
        GenTranslate  (fine-tuned) & 21.4 & 30.7 & 26.1 \\
        \rowcolor{yellow!20} \method-Gemma-MoE (zero-shot)& 18.1 & 27.6 & 22.9 \\
        \bottomrule
    \end{tabular}
    }
    \label{tab:mt_results}
\end{table}

To further assess the generalization ability of \methodwithspace, we evaluate it on the WMT'20 machine translation benchmark for Japanese and Chinese in a zero-shot setting. As shown in Table \ref{tab:mt_results}, \methodwithspace achieves competitive performance compared to fine-tuned MT models, obtaining an average BLEU score of 22.9. This result highlights the potential of \methodwithspace to handle unseen translation tasks by leveraging the knowledge learned from pre-training.

\paragraph{OCR and TEC} For the OCR task, \methodwithspace achieves a substantial error reduction, lowering the WER from 71.03\% to 14.43\%. This represents a significant improvement over the baseline and demonstrates the model's ability to correct OCR errors effectively. Compared to the Mixtral-MoE model fine-tuned directly on the OCR dataset, \methodwithspace obtains a 1.02\% lower WER, highlighting the benefit of the task-guided expert assignment approach. In the TEC task, \methodwithspace showcases its versatility by improving the performance on both grammar correction and coherence improvement subtasks. For grammar correction, \methodwithspace reduces the WER from 31.41\% to 9.42\%, outperforming the directly fine-tuned Mixtral-MoE model by 1.31\%. On the coherence subtask, \methodwithspace achieves a WER of 9.71\%, which is 0.46\% higher than the directly fine-tuned model but still a significant improvement over the baseline.
\vspace{0.2cm}
\begin{table}[ht!]
    \centering
    \caption{WER comparison of \methodwithspace against the baseline and a directly fine-tuned Mixtral-MoE model  (8x7B) on grammar correction and coherence improvement tasks from the CoEdIT dataset \citep{raheja2023coedit}, and the OCR task using the PleIAs/Post-OCR-Correction dataset \citep{PleIAs_Post_OCR_Correction}. }
    \resizebox{1\columnwidth}{!}{
    \begin{tabular}{lcccc}
        \toprule
        Task / WER $\downarrow$& Grammar Correction & Coherence Improv. & OCR \\
        \midrule
        Mixtral-MoE (frozen) & 31.41 & 13.48 & 71.03 \\
        GPT-3.5-turbo & 17.43 & ~~12.25 & 39.45 \\
        Mixtral-MoE-FFT & 10.73 & ~~12.05 & 45.32 \\
        \rowcolor{yellow!20}\method-Mixtral-MoE & ~~9.42 & ~~9.71 & 14.43 \\
        \bottomrule
    \end{tabular}}
    \label{tab:ocr_gec_results}
\end{table}


\vspace{0.3cm}
\section{Conclusion}
In this work, we proposed \method, a multi-task GER approach that leverages task-guided MoEs to handle diverse tasks. \methodwithspace assigns each expert to a specific dataset during training, enabling the experts to capture task-specific features while allowing knowledge sharing through the gating network. Our results show that task-guided expert assignment is a promising approach for multi-task learning in error correction and other natural language processing tasks. By aligning experts with datasets, \methodwithspace can effectively capture the nuances and specificities of each task while benefiting from the shared knowledge learned by the gating network and other model components. Future work includes exploring more advanced expert assignment strategies, such as dynamically assigning experts based on the input characteristics. 




\clearpage
\section*{Ethical Considerations}

We aim to provide a transparent and comprehensive understanding of the current scope of \textsc{NeKo}, and pave the way for future research to further improve the \textsc{NeKo} model.
\paragraph{Dataset Diversity and Size and Assumptions in Error Distribution}
This study addresses a mixture of error correction tasks, including ASR, ST, OCR, and TEC, using representative task-specific datasets such as LibriSpeech for ASR, CoVoST for ST, ICDAR 2019 for OCR, and CoNLL-2014 for TEC. While these datasets are widely recognized benchmarks, they may not cover all possible error correction scenarios, particularly those involving more complex or less common error types found in real-world data. This setup assumes that the error distributions in the training datasets are representative of those in real-world applications. Consequently, the performance of \method might be overestimated for certain types of data not covered by these benchmarks, affecting the generalizability of the results to more diverse and noisy real-world scenarios. Future research should include a broader range of datasets, particularly those with more diverse and challenging error types, and investigate methods to dynamically adapt to varying error distributions, possibly through online learning~\cite{yasunaga2021lm} or domain adaptation techniques~\cite{khurana2021unsupervised}, to better evaluate the robustness and generalizability of the model.

\paragraph{Societal Considerations}
The study does not extensively address the ethical and societal implications of deploying \textsc{NeKo} in real-world applications. There could be unintended consequences, such as biases in error correction or misuse of the technology in sensitive applications. Future work should include a thorough analysis of the ethical and societal impacts of the model, along with strategies to mitigate potential negative consequences. This could involve incorporating fairness and bias detection mechanisms~\cite{liu2022quantifying} into the model to ensure responsible and ethical deployment.

\paragraph{Boarder Impacts} The \textsc{NeKo} model's application of MoE for multi-domain and multi-task error correction has the potential to significantly enhance automated system's performance across various domains, such as healthcare, education and customer service. By improving standard mediums of communication such as speech recognition, translation and optical character recognition \textsc{NeKo} can facilitate more inclusive technologies, benefiting individuals with impairments or non-native speakers. Additionally, the economic benefits from reduced manual correction efforts and educational advantages from more accurate communication system can be substantial. The open-sourcing of \textsc{NeKo} under the CC BY-SA 4.0 license encourages collaboration and reproducibility with in the reserach community, fostering innovation and broader application. Future work should also consider optimizing the training process to minimize the environmental impact, promoting sustainable AI development practices.

\bibliography{custom, esb,latex/ref}

\begin{thebibliography}{74}
\providecommand{\natexlab}[1]{#1}

\bibitem[{Ackley et~al.(1985)Ackley, Hinton, and Sejnowski}]{DBLP:journals/cogsci/AckleyHS85}
David~H. Ackley, Geoffrey~E. Hinton, and Terrence~J. Sejnowski. 1985.
\newblock \href {https://doi.org/10.1207/S15516709COG0901\_7} {A learning algorithm for boltzmann machines}.
\newblock \emph{Cogn. Sci.}, 9(1):147--169.

\bibitem[{Aks{\"e}nova et~al.(2021)Aks{\"e}nova, van Esch, Flynn, and Golik}]{aksenova2021might}
Al{\"e}na Aks{\"e}nova, Daan van Esch, James Flynn, and Pavel Golik. 2021.
\newblock How might we create better benchmarks for speech recognition?
\newblock In \emph{Proceedings of the 1st workshop on benchmarking: Past, present and future}, pages 22--34.

\bibitem[{Ardila et~al.(2020)Ardila, Branson, Davis, Kohler, Meyer, Henretty, Morais, Saunders, Tyers, and Weber}]{DBLP:conf/lrec/ArdilaBDKMHMSTW20}
Rosana Ardila, Megan Branson, Kelly Davis, Michael Kohler, Josh Meyer, Michael Henretty, Reuben Morais, Lindsay Saunders, Francis~M. Tyers, and Gregor Weber. 2020.
\newblock \href {https://aclanthology.org/2020.lrec-1.520/} {Common voice: {A} massively-multilingual speech corpus}.
\newblock In \emph{Proceedings of The 12th Language Resources and Evaluation Conference, {LREC} 2020, Marseille, France, May 11-16, 2020}, pages 4218--4222. European Language Resources Association.

\bibitem[{Barrault et~al.(2020{\natexlab{a}})Barrault, Biesialska, Bojar, Costa{-}juss{\`{a}}, Federmann, Graham, Grundkiewicz, Haddow, Huck, Joanis, Kocmi, Koehn, Lo, Ljubesic, Monz, Morishita, Nagata, Nakazawa, Pal, Post, and Zampieri}]{DBLP:conf/wmt/BarraultBBCFGGH20}
Lo{\"{\i}}c Barrault, Magdalena Biesialska, Ondrej Bojar, Marta~R. Costa{-}juss{\`{a}}, Christian Federmann, Yvette Graham, Roman Grundkiewicz, Barry Haddow, Matthias Huck, Eric Joanis, Tom Kocmi, Philipp Koehn, Chi{-}kiu Lo, Nikola Ljubesic, Christof Monz, Makoto Morishita, Masaaki Nagata, Toshiaki Nakazawa, Santanu Pal, Matt Post, and Marcos Zampieri. 2020{\natexlab{a}}.
\newblock \href {https://aclanthology.org/2020.wmt-1.1/} {Findings of the 2020 conference on machine translation {(WMT20)}}.
\newblock In \emph{Proceedings of the Fifth Conference on Machine Translation, WMT@EMNLP 2020, Online, November 19-20, 2020}, pages 1--55. Association for Computational Linguistics.

\bibitem[{Barrault et~al.(2020{\natexlab{b}})Barrault, Biesialska, Bojar, Costa-juss{\`a}, Federmann, Graham, Grundkiewicz, Haddow, Huck, Joanis, Kocmi, Koehn, Lo, Ljube{\v{s}}i{\'c}, Monz, Morishita, Nagata, Nakazawa, Pal, Post, and Zampieri}]{barrault-etal-2020-findings}
Lo{\"\i}c Barrault, Magdalena Biesialska, Ond{\v{r}}ej Bojar, Marta~R. Costa-juss{\`a}, Christian Federmann, Yvette Graham, Roman Grundkiewicz, Barry Haddow, Matthias Huck, Eric Joanis, Tom Kocmi, Philipp Koehn, Chi-kiu Lo, Nikola Ljube{\v{s}}i{\'c}, Christof Monz, Makoto Morishita, Masaaki Nagata, Toshiaki Nakazawa, Santanu Pal, Matt Post, and Marcos Zampieri. 2020{\natexlab{b}}.
\newblock \href {https://aclanthology.org/2020.wmt-1.1} {Findings of the 2020 conference on machine translation ({WMT}20)}.
\newblock In \emph{Proceedings of the Fifth Conference on Machine Translation}, pages 1--55, Online. Association for Computational Linguistics.

\bibitem[{Barrault et~al.(2023{\natexlab{a}})Barrault, Chung, Meglioli, Dale, Dong, Duquenne, Elsahar, Gong, Heffernan, Hoffman et~al.}]{barrault2023seamlessm4t}
Lo{\"\i}c Barrault, Yu-An Chung, Mariano~Cora Meglioli, David Dale, Ning Dong, Paul-Ambroise Duquenne, Hady Elsahar, Hongyu Gong, Kevin Heffernan, John Hoffman, et~al. 2023{\natexlab{a}}.
\newblock Seamlessm4t-massively multilingual \& multimodal machine translation.
\newblock \emph{arXiv preprint arXiv:2308.11596}.

\bibitem[{Barrault et~al.(2023{\natexlab{b}})Barrault, Chung, Meglioli, Dale, Dong, Duppenthaler, Duquenne, Ellis, Elsahar, Haaheim et~al.}]{barrault2023seamlessv2}
Lo{\"\i}c Barrault, Yu-An Chung, Mariano~Coria Meglioli, David Dale, Ning Dong, Mark Duppenthaler, Paul-Ambroise Duquenne, Brian Ellis, Hady Elsahar, Justin Haaheim, et~al. 2023{\natexlab{b}}.
\newblock Seamless: Multilingual expressive and streaming speech translation.
\newblock \emph{arXiv preprint arXiv:2312.05187}.

\bibitem[{Brown et~al.(2020)Brown, Mann, Ryder, Subbiah, Kaplan, Dhariwal, Neelakantan, Shyam, Sastry, Askell, Agarwal, Herbert-Voss, Krueger, Henighan, Child, Ramesh, Ziegler, Wu, Winter, Hesse, Chen, Sigler, Litwin, Gray, Chess, Clark, Berner, McCandlish, Radford, Sutskever, and Amodei}]{Brown2020LanguageMA}
Tom~B. Brown, Benjamin Mann, Nick Ryder, Melanie Subbiah, Jared Kaplan, Prafulla Dhariwal, Arvind Neelakantan, Pranav Shyam, Girish Sastry, Amanda Askell, Sandhini Agarwal, Ariel Herbert-Voss, Gretchen Krueger, T.~J. Henighan, Rewon Child, Aditya Ramesh, Daniel~M. Ziegler, Jeff Wu, Clemens Winter, Christopher Hesse, Mark Chen, Eric Sigler, Mateusz Litwin, Scott Gray, Benjamin Chess, Jack Clark, Christopher Berner, Sam McCandlish, Alec Radford, Ilya Sutskever, and Dario Amodei. 2020.
\newblock \href {https://api.semanticscholar.org/CorpusID:218971783} {Language models are few-shot learners}.
\newblock \emph{ArXiv}, abs/2005.14165.

\bibitem[{Carletta(2007)}]{carletta07_ami}
Jean Carletta. 2007.
\newblock \href {https://doi.org/10.1007/s10579-007-9040-x} {{Unleashing the killer corpus: experiences in creating the multi-everything AMI Meeting Corpus}}.
\newblock \emph{Language Resources and Evaluation}, 41(2):181--190.

\bibitem[{Chan et~al.(2023)Chan, Myers, Vijayanarasimhan, Ross, and Canny}]{chan2023ic3}
David Chan, Austin Myers, Sudheendra Vijayanarasimhan, David Ross, and John Canny. 2023.
\newblock Ic3: Image captioning by committee consensus.
\newblock In \emph{Proceedings of the 2023 Conference on Empirical Methods in Natural Language Processing}, pages 8975--9003.

\bibitem[{Chan et~al.(2016)Chan, Jaitly, Le, and Vinyals}]{chan2016listen}
William Chan, Navdeep Jaitly, Quoc Le, and Oriol Vinyals. 2016.
\newblock Listen, attend and spell: A neural network for large vocabulary conversational speech recognition.
\newblock In \emph{2016 IEEE international conference on acoustics, speech and signal processing (ICASSP)}, pages 4960--4964. IEEE.

\bibitem[{CHEN et~al.(2023)CHEN, Hu, Yang, Siniscalchi, Chen, and Chng}]{chen2023hyporadise}
CHEN CHEN, Yuchen Hu, Chao-Han~Huck Yang, Sabato~Marco Siniscalchi, Pin-Yu Chen, and Ensiong Chng. 2023.
\newblock Hyporadise: An open baseline for generative speech recognition with large language models.
\newblock In \emph{Thirty-seventh Conference on Neural Information Processing Systems Datasets and Benchmarks Track}.

\bibitem[{Chen et~al.(2024{\natexlab{a}})Chen, Li, Hu, Siniscalchi, Chen, Chng, and Yang}]{chen2024s}
Chen Chen, Ruizhe Li, Yuchen Hu, Sabato~Marco Siniscalchi, Pin-Yu Chen, Ensiong Chng, and Chao-Han~Huck Yang. 2024{\natexlab{a}}.
\newblock It's never too late: Fusing acoustic information into large language models for automatic speech recognition.
\newblock \emph{arXiv preprint arXiv:2402.05457}.

\bibitem[{{Chen} et~al.(2021){Chen}, {Chai}, {Wang}, {Du}, {Zhang}, {Weng}, {Su}, {Povey}, {Trmal}, {Zhang}, {Jin}, {Khudanpur}, {Watanabe}, {Zhao}, {Zou}, {Li}, {Yao}, {Wang}, {Wang}, {You}, and {Yan}}]{chen21_gigaspeech}
Guoguo {Chen}, Shuzhou {Chai}, Guanbo {Wang}, Jiayu {Du}, Wei-Qiang {Zhang}, Chao {Weng}, Dan {Su}, Daniel {Povey}, Jan {Trmal}, Junbo {Zhang}, Mingjie {Jin}, Sanjeev {Khudanpur}, Shinji {Watanabe}, Shuaijiang {Zhao}, Wei {Zou}, Xiangang {Li}, Xuchen {Yao}, Yongqing {Wang}, Yujun {Wang}, Zhao {You}, and Zhiyong {Yan}. 2021.
\newblock \href {https://arxiv.org/abs/2106.06909} {{GigaSpeech: An Evolving, Multi-domain ASR Corpus with 10,000 Hours of Transcribed Audio}}.
\newblock \emph{arXiv e-prints}, arXiv:2106.06909.

\bibitem[{Chen et~al.(2024{\natexlab{b}})Chen, Huang, Andrusenko, Hrinchuk, Puvvada, Li, Ghosh, Balam, and Ginsburg}]{chen2024salm}
Zhehuai Chen, He~Huang, Andrei Andrusenko, Oleksii Hrinchuk, Krishna~C Puvvada, Jason Li, Subhankar Ghosh, Jagadeesh Balam, and Boris Ginsburg. 2024{\natexlab{b}}.
\newblock Salm: Speech-augmented language model with in-context learning for speech recognition and translation.
\newblock In \emph{ICASSP 2024-2024 IEEE International Conference on Acoustics, Speech and Signal Processing (ICASSP)}, pages 13521--13525. IEEE.

\bibitem[{Chen et~al.(2024{\natexlab{c}})Chen, Huang, Hrinchuk, Puvvada, Koluguri, {\.Z}elasko, Balam, and Ginsburg}]{chen2024bestow}
Zhehuai Chen, He~Huang, Oleksii Hrinchuk, Krishna~C Puvvada, Nithin~Rao Koluguri, Piotr {\.Z}elasko, Jagadeesh Balam, and Boris Ginsburg. 2024{\natexlab{c}}.
\newblock Bestow: Efficient and streamable speech language model with the best of two worlds in gpt and t5.
\newblock \emph{arXiv preprint arXiv:2406.19954}.

\bibitem[{Cheng et~al.(2021)Cheng, Burgess, Vernooij, Sol{\'\i}s-Barroso, McDermott, and Namboodiripad}]{cheng2021problematic}
Lauretta~SP Cheng, Danielle Burgess, Natasha Vernooij, Cecilia Sol{\'\i}s-Barroso, Ashley McDermott, and Savithry Namboodiripad. 2021.
\newblock The problematic concept of native speaker in psycholinguistics: Replacing vague and harmful terminology with inclusive and accurate measures.
\newblock \emph{Frontiers in psychology}, 12:715843.

\bibitem[{Chu et~al.(2024)Chu, Xu, Yang, Wei, Wei, Guo, Leng, Lv, He, Lin et~al.}]{chu2024qwen2}
Yunfei Chu, Jin Xu, Qian Yang, Haojie Wei, Xipin Wei, Zhifang Guo, Yichong Leng, Yuanjun Lv, Jinzheng He, Junyang Lin, et~al. 2024.
\newblock Qwen2-audio technical report.
\newblock \emph{arXiv preprint arXiv:2407.10759}.

\bibitem[{Conneau et~al.(2022)Conneau, Ma, Khanuja, Zhang, Axelrod, Dalmia, Riesa, Rivera, and Bapna}]{DBLP:conf/slt/ConneauMKZADRRB22}
Alexis Conneau, Min Ma, Simran Khanuja, Yu~Zhang, Vera Axelrod, Siddharth Dalmia, Jason Riesa, Clara Rivera, and Ankur Bapna. 2022.
\newblock \href {https://doi.org/10.1109/SLT54892.2023.10023141} {{FLEURS:} few-shot learning evaluation of universal representations of speech}.
\newblock In \emph{{IEEE} Spoken Language Technology Workshop, {SLT} 2022, Doha, Qatar, January 9-12, 2023}, pages 798--805. {IEEE}.

\bibitem[{Costa{-}juss{\`{a}} et~al.(2022)Costa{-}juss{\`{a}}, Cross, {\c{C}}elebi, Elbayad, Heafield, Heffernan, Kalbassi, Lam, Licht, Maillard, Sun, Wang, Wenzek, Youngblood, Akula, Barrault, Gonzalez, Hansanti, Hoffman, Jarrett, Sadagopan, Rowe, Spruit, Tran, Andrews, Ayan, Bhosale, Edunov, Fan, Gao, Goswami, Guzm{\'{a}}n, Koehn, Mourachko, Ropers, Saleem, Schwenk, and Wang}]{DBLP:journals/corr/abs-2207-04672}
Marta~R. Costa{-}juss{\`{a}}, James Cross, Onur {\c{C}}elebi, Maha Elbayad, Kenneth Heafield, Kevin Heffernan, Elahe Kalbassi, Janice Lam, Daniel Licht, Jean Maillard, Anna~Y. Sun, Skyler Wang, Guillaume Wenzek, Al~Youngblood, Bapi Akula, Lo{\"{\i}}c Barrault, Gabriel~Mejia Gonzalez, Prangthip Hansanti, John Hoffman, Semarley Jarrett, Kaushik~Ram Sadagopan, Dirk Rowe, Shannon Spruit, Chau Tran, Pierre Andrews, Necip~Fazil Ayan, Shruti Bhosale, Sergey Edunov, Angela Fan, Cynthia Gao, Vedanuj Goswami, Francisco Guzm{\'{a}}n, Philipp Koehn, Alexandre Mourachko, Christophe Ropers, Safiyyah Saleem, Holger Schwenk, and Jeff Wang. 2022.
\newblock \href {https://doi.org/10.48550/ARXIV.2207.04672} {No language left behind: Scaling human-centered machine translation}.
\newblock \emph{CoRR}, abs/2207.04672.

\bibitem[{Dai et~al.(2024)Dai, Deng, Zhao, Xu, Gao, Chen, Li, Zeng, Yu, Wu, Xie, Li, Huang, Luo, Ruan, Sui, and Liang}]{DBLP:journals/corr/abs-2401-06066}
Damai Dai, Chengqi Deng, Chenggang Zhao, R.~X. Xu, Huazuo Gao, Deli Chen, Jiashi Li, Wangding Zeng, Xingkai Yu, Y.~Wu, Zhenda Xie, Y.~K. Li, Panpan Huang, Fuli Luo, Chong Ruan, Zhifang Sui, and Wenfeng Liang. 2024.
\newblock \href {https://doi.org/10.48550/ARXIV.2401.06066} {Deepseekmoe: Towards ultimate expert specialization in mixture-of-experts language models}.
\newblock \emph{CoRR}, abs/2401.06066.

\bibitem[{{Del Rio} et~al.(2022){Del Rio}, {Ha}, {McNamara}, {Miller}, and {Chandra}}]{delrio22_earnings22}
Miguel {Del Rio}, Peter {Ha}, Quinten {McNamara}, Corey {Miller}, and Shipra {Chandra}. 2022.
\newblock \href {https://arxiv.org/abs/2203.15591} {{Earnings-22: A Practical Benchmark for Accents in the Wild}}.
\newblock \emph{arXiv e-prints}, arXiv:2203.15591.

\bibitem[{Deng et~al.(2013)Deng, Li, Huang, Yao, Yu, Seide, Seltzer, Zweig, He, Williams et~al.}]{deng2013recent}
Li~Deng, Jinyu Li, Jui-Ting Huang, Kaisheng Yao, Dong Yu, Frank Seide, Michael Seltzer, Geoff Zweig, Xiaodong He, Jason Williams, et~al. 2013.
\newblock Recent advances in deep learning for speech research at microsoft.
\newblock In \emph{2013 IEEE international conference on acoustics, speech and signal processing}, pages 8604--8608. IEEE.

\bibitem[{Di~Gangi et~al.(2019)Di~Gangi, Cattoni, Bentivogli, Negri, and Turchi}]{di-gangi-etal-2019-must}
Mattia~A. Di~Gangi, Roldano Cattoni, Luisa Bentivogli, Matteo Negri, and Marco Turchi. 2019.
\newblock \href {https://doi.org/10.18653/v1/N19-1202} {{M}u{ST}-{C}: a {M}ultilingual {S}peech {T}ranslation {C}orpus}.
\newblock In \emph{Proceedings of the 2019 Conference of the North {A}merican Chapter of the Association for Computational Linguistics: Human Language Technologies, Volume 1 (Long and Short Papers)}, pages 2012--2017, Minneapolis, Minnesota. Association for Computational Linguistics.

\bibitem[{Du et~al.(2016)Du, Tu, Sun, Ma, Wang, Pan, Liu, Chen, and Lee}]{du2016ustc}
Jun Du, Yan-Hui Tu, Lei Sun, Feng Ma, Hai-Kun Wang, Jia Pan, Cong Liu, Jing-Dong Chen, and Chin-Hui Lee. 2016.
\newblock The ustc-iflytek system for chime-4 challenge.
\newblock \emph{Proc. CHiME}, 4(1):36--38.

\bibitem[{Fedus et~al.(2022)Fedus, Zoph, and Shazeer}]{DBLP:journals/jmlr/FedusZS22}
William Fedus, Barret Zoph, and Noam Shazeer. 2022.
\newblock \href {http://jmlr.org/papers/v23/21-0998.html} {Switch transformers: Scaling to trillion parameter models with simple and efficient sparsity}.
\newblock \emph{J. Mach. Learn. Res.}, 23:120:1--120:39.

\bibitem[{Gandhi et~al.(2022)Gandhi, von Platen, and Rush}]{DBLP:journals/corr/abs-2210-13352}
Sanchit Gandhi, Patrick von Platen, and Alexander~M. Rush. 2022.
\newblock \href {https://doi.org/10.48550/ARXIV.2210.13352} {{ESB:} {A} benchmark for multi-domain end-to-end speech recognition}.
\newblock \emph{CoRR}, abs/2210.13352.

\bibitem[{Gandhi et~al.(2023)Gandhi, von Platen, and Rush}]{gandhi2023distil}
Sanchit Gandhi, Patrick von Platen, and Alexander~M Rush. 2023.
\newblock Distil-whisper: Robust knowledge distillation via large-scale pseudo labelling.
\newblock \emph{arXiv preprint arXiv:2311.00430}.

\bibitem[{He et~al.(2019)He, Sainath, Prabhavalkar, McGraw, Alvarez, Zhao, Rybach, Kannan, Wu, Pang et~al.}]{he2019streaming}
Yanzhang He, Tara~N Sainath, Rohit Prabhavalkar, Ian McGraw, Raziel Alvarez, Ding Zhao, David Rybach, Anjuli Kannan, Yonghui Wu, Ruoming Pang, et~al. 2019.
\newblock Streaming end-to-end speech recognition for mobile devices.
\newblock In \emph{ICASSP 2019-2019 IEEE International Conference on Acoustics, Speech and Signal Processing (ICASSP)}, pages 6381--6385. IEEE.

\bibitem[{Hernandez et~al.(2018)Hernandez, Nguyen, Ghannay, Tomashenko, and Est{\`e}ve}]{hernandez18_tedlium}
Fran{\c{c}}ois Hernandez, Vincent Nguyen, Sahar Ghannay, Natalia Tomashenko, and Yannick Est{\`e}ve. 2018.
\newblock {TED-LIUM 3: Twice as Much Data and Corpus Repartition for Experiments on Speaker Adaptation}.
\newblock In \emph{Speech and Computer}, pages 198--208. Springer International Publishing.

\bibitem[{Hu et~al.(2021)Hu, Shen, Wallis, Allen-Zhu, Li, Wang, Wang, and Chen}]{hu2021lora}
Edward~J Hu, Yelong Shen, Phillip Wallis, Zeyuan Allen-Zhu, Yuanzhi Li, Shean Wang, Lu~Wang, and Weizhu Chen. 2021.
\newblock Lora: Low-rank adaptation of large language models.
\newblock \emph{arXiv preprint arXiv:2106.09685}.

\bibitem[{Hu et~al.(2024{\natexlab{a}})Hu, Chen, Yang, Li, Zhang, Chen, and Chng}]{hu2024large}
Yuchen Hu, Chen Chen, Chao-Han~Huck Yang, Ruizhe Li, Chao Zhang, Pin-Yu Chen, and EnSiong Chng. 2024{\natexlab{a}}.
\newblock Large language models are efficient learners of noise-robust speech recognition.
\newblock \emph{arXiv preprint arXiv:2401.10446}.

\bibitem[{Hu et~al.(2024{\natexlab{b}})Hu, Chen, Yang, Li, Zhang, Chen, and Chng}]{DBLP:journals/corr/abs-2402-06894}
Yuchen Hu, Chen Chen, Chao{-}Han~Huck Yang, Ruizhe Li, Dong Zhang, Zhehuai Chen, and Eng~Siong Chng. 2024{\natexlab{b}}.
\newblock \href {https://doi.org/10.48550/ARXIV.2402.06894} {Gentranslate: Large language models are generative multilingual speech and machine translators}.
\newblock \emph{CoRR}, abs/2402.06894.

\bibitem[{Hu et~al.(2024{\natexlab{c}})Hu, Chen, Yang, Li, Zhang, Chen, and Chng}]{hu2024gentranslate}
Yuchen Hu, Chen Chen, Chao-Han~Huck Yang, Ruizhe Li, Dong Zhang, Zhehuai Chen, and Eng~Siong Chng. 2024{\natexlab{c}}.
\newblock Gentranslate: Large language models are generative multilingual speech and machine translators.
\newblock \emph{arXiv preprint arXiv:2402.06894}.

\bibitem[{Jiang et~al.(2023)Jiang, Sablayrolles, Mensch, Bamford, Chaplot, Casas, Bressand, Lengyel, Lample, Saulnier et~al.}]{jiang2023mistral}
Albert~Q Jiang, Alexandre Sablayrolles, Arthur Mensch, Chris Bamford, Devendra~Singh Chaplot, Diego de~las Casas, Florian Bressand, Gianna Lengyel, Guillaume Lample, Lucile Saulnier, et~al. 2023.
\newblock Mistral 7b.
\newblock \emph{arXiv preprint arXiv:2310.06825}.

\bibitem[{Jiang et~al.(2024{\natexlab{a}})Jiang, Sablayrolles, Roux, Mensch, Savary, Bamford, Chaplot, de~Las~Casas, Hanna, Bressand, Lengyel, Bour, Lample, Lavaud, Saulnier, Lachaux, Stock, Subramanian, Yang, Antoniak, Scao, Gervet, Lavril, Wang, Lacroix, and Sayed}]{DBLP:journals/corr/abs-2401-04088}
Albert~Q. Jiang, Alexandre Sablayrolles, Antoine Roux, Arthur Mensch, Blanche Savary, Chris Bamford, Devendra~Singh Chaplot, Diego de~Las~Casas, Emma~Bou Hanna, Florian Bressand, Gianna Lengyel, Guillaume Bour, Guillaume Lample, L{\'{e}}lio~Renard Lavaud, Lucile Saulnier, Marie{-}Anne Lachaux, Pierre Stock, Sandeep Subramanian, Sophia Yang, Szymon Antoniak, Teven~Le Scao, Th{\'{e}}ophile Gervet, Thibaut Lavril, Thomas Wang, Timoth{\'{e}}e Lacroix, and William~El Sayed. 2024{\natexlab{a}}.
\newblock \href {https://doi.org/10.48550/ARXIV.2401.04088} {Mixtral of experts}.
\newblock \emph{CoRR}, abs/2401.04088.

\bibitem[{Jiang et~al.(2024{\natexlab{b}})Jiang, Sablayrolles, Roux, Mensch, Savary, Bamford, Chaplot, de~Las~Casas, Hanna, Bressand, Lengyel, Bour, Lample, Lavaud, Saulnier, Lachaux, Stock, Subramanian, Yang, Antoniak, Scao, Gervet, Lavril, Wang, Lacroix, and Sayed}]{Jiang2024MixtralOE}
Albert~Q. Jiang, Alexandre Sablayrolles, Antoine Roux, Arthur Mensch, Blanche Savary, Chris Bamford, Devendra~Singh Chaplot, Diego de~Las~Casas, Emma~Bou Hanna, Florian Bressand, Gianna Lengyel, Guillaume Bour, Guillaume Lample, L'elio~Renard Lavaud, Lucile Saulnier, Marie-Anne Lachaux, Pierre Stock, Sandeep Subramanian, Sophia Yang, Szymon Antoniak, Teven~Le Scao, Th{\'e}ophile Gervet, Thibaut Lavril, Thomas Wang, Timoth{\'e}e Lacroix, and William~El Sayed. 2024{\natexlab{b}}.
\newblock \href {https://api.semanticscholar.org/CorpusID:266844877} {Mixtral of experts}.
\newblock \emph{ArXiv}, abs/2401.04088.

\bibitem[{Jiang et~al.(2020)Jiang, Gossack-Keenan, and Pell}]{jiang2020believe}
Xiaoming Jiang, Kira Gossack-Keenan, and Marc~D Pell. 2020.
\newblock To believe or not to believe? how voice and accent information in speech alter listener impressions of trust.
\newblock \emph{Quarterly Journal of Experimental Psychology}, 73(1):55--79.

\bibitem[{Khurana et~al.(2021)Khurana, Moritz, Hori, and Le~Roux}]{khurana2021unsupervised}
Sameer Khurana, Niko Moritz, Takaaki Hori, and Jonathan Le~Roux. 2021.
\newblock Unsupervised domain adaptation for speech recognition via uncertainty driven self-training.
\newblock In \emph{ICASSP 2021-2021 IEEE International Conference on Acoustics, Speech and Signal Processing (ICASSP)}, pages 6553--6557. IEEE.

\bibitem[{Komatsuzaki et~al.(2023)Komatsuzaki, Puigcerver, Lee{-}Thorp, Ruiz, Mustafa, Ainslie, Tay, Dehghani, and Houlsby}]{DBLP:conf/iclr/KomatsuzakiPLRM23}
Aran Komatsuzaki, Joan Puigcerver, James Lee{-}Thorp, Carlos~Riquelme Ruiz, Basil Mustafa, Joshua Ainslie, Yi~Tay, Mostafa Dehghani, and Neil Houlsby. 2023.
\newblock \href {https://openreview.net/pdf?id=T5nUQDrM4u} {Sparse upcycling: Training mixture-of-experts from dense checkpoints}.
\newblock In \emph{The Eleventh International Conference on Learning Representations, {ICLR} 2023, Kigali, Rwanda, May 1-5, 2023}. OpenReview.net.

\bibitem[{Lepikhin et~al.(2021)Lepikhin, Lee, Xu, Chen, Firat, Huang, Krikun, Shazeer, and Chen}]{DBLP:conf/iclr/LepikhinLXCFHKS21}
Dmitry Lepikhin, HyoukJoong Lee, Yuanzhong Xu, Dehao Chen, Orhan Firat, Yanping Huang, Maxim Krikun, Noam Shazeer, and Zhifeng Chen. 2021.
\newblock \href {https://openreview.net/forum?id=qrwe7XHTmYb} {Gshard: Scaling giant models with conditional computation and automatic sharding}.
\newblock In \emph{9th International Conference on Learning Representations, {ICLR} 2021, Virtual Event, Austria, May 3-7, 2021}. OpenReview.net.

\bibitem[{Lev-Ari(2015)}]{lev2015comprehending}
Shiri Lev-Ari. 2015.
\newblock Comprehending non-native speakers: Theory and evidence for adjustment in manner of processing.
\newblock \emph{Frontiers in psychology}, 5:111794.

\bibitem[{Levinson and Evans(2010)}]{levinson2010time}
Stephen~C Levinson and Nicholas Evans. 2010.
\newblock Time for a sea-change in linguistics: Response to comments on ‘the myth of language universals’.
\newblock \emph{Lingua}, 120(12):2733--2758.

\bibitem[{Liu et~al.(2022)Liu, Jia, Wei, Xu, and Vosoughi}]{liu2022quantifying}
Ruibo Liu, Chenyan Jia, Jason Wei, Guangxuan Xu, and Soroush Vosoughi. 2022.
\newblock Quantifying and alleviating political bias in language models.
\newblock \emph{Artificial Intelligence}, 304:103654.

\bibitem[{Loshchilov and Hutter(2019)}]{DBLP:conf/iclr/LoshchilovH19}
Ilya Loshchilov and Frank Hutter. 2019.
\newblock \href {https://openreview.net/forum?id=Bkg6RiCqY7} {Decoupled weight decay regularization}.
\newblock In \emph{7th International Conference on Learning Representations, {ICLR} 2019, New Orleans, LA, USA, May 6-9, 2019}. OpenReview.net.

\bibitem[{Marshall et~al.(2015)Marshall, Jones, Denmark, Mason, Atkinson, Botting, and Morgan}]{marshall2015deaf}
Chlo{\"e} Marshall, Anna Jones, Tanya Denmark, Kathryn Mason, Joanna Atkinson, Nicola Botting, and Gary Morgan. 2015.
\newblock Deaf children's non-verbal working memory is impacted by their language experience.
\newblock \emph{Frontiers in psychology}, 6:527.

\bibitem[{NVIDIA(2024)}]{canary}
NVIDIA. 2024.
\newblock \href {https://developer.nvidia.com/blog/new-standard-for-speech-recognition-and-translation-from-the-nvidia-nemo-canary-model/} {New standard for speech recognition and translation from the nvidia nemo canary model}.
\newblock Accessed: 2024-05-20.

\bibitem[{O’Neill et~al.(2021)O’Neill, Lavrukhin, Majumdar, Noroozi, Zhang, Kuchaiev, Balam, Dovzhenko, Freyberg, Shulman, Ginsburg, Watanabe, and Kucsko}]{oneill21_kensho}
Patrick~K. O’Neill, Vitaly Lavrukhin, Somshubra Majumdar, Vahid Noroozi, Yuekai Zhang, Oleksii Kuchaiev, Jagadeesh Balam, Yuliya Dovzhenko, Keenan Freyberg, Michael~D. Shulman, Boris Ginsburg, Shinji Watanabe, and Georg Kucsko. 2021.
\newblock \href {https://doi.org/10.21437/Interspeech.2021-1860} {{SPGISpeech: 5,000 Hours of Transcribed Financial Audio for Fully Formatted End-to-End Speech Recognition}}.
\newblock In \emph{Proc. Interspeech 2021}, pages 1434--1438.

\bibitem[{Panayotov et~al.(2015)Panayotov, Chen, Povey, and Khudanpur}]{panayotov15_libripseech}
Vassil Panayotov, Guoguo Chen, Daniel Povey, and Sanjeev Khudanpur. 2015.
\newblock \href {https://doi.org/10.1109/ICASSP.2015.7178964} {{Librispeech: An ASR corpus based on public domain audio books}}.
\newblock In \emph{2015 IEEE International Conference on Acoustics, Speech and Signal Processing (ICASSP)}, pages 5206--5210.

\bibitem[{Papineni et~al.(2002)Papineni, Roukos, Ward, and Zhu}]{DBLP:conf/acl/PapineniRWZ02}
Kishore Papineni, Salim Roukos, Todd Ward, and Wei{-}Jing Zhu. 2002.
\newblock \href {https://doi.org/10.3115/1073083.1073135} {Bleu: a method for automatic evaluation of machine translation}.
\newblock In \emph{Proceedings of the 40th Annual Meeting of the Association for Computational Linguistics, July 6-12, 2002, Philadelphia, PA, {USA}}, pages 311--318. {ACL}.

\bibitem[{PleIAs(2023)}]{PleIAs_Post_OCR_Correction}
PleIAs. 2023.
\newblock \href {https://huggingface.co/datasets/PleIAs/Post-OCR-Correction} {Post-ocr-correction}.

\bibitem[{Radford et~al.(2022)Radford, Kim, Xu, Brockman, McLeavey, and Sutskever}]{radford22_whipser}
Alec Radford, Jong~Wook Kim, Tao Xu, Greg Brockman, Christine McLeavey, and Ilya Sutskever. 2022.
\newblock \href {https://cdn.openai.com/papers/whisper.pdf} {{Robust Speech Recognition via Large-Scale Weak Supervision}}.
\newblock Technical report, OpenAI.

\bibitem[{Raffel et~al.(2020)Raffel, Shazeer, Roberts, Lee, Narang, Matena, Zhou, Li, and Liu}]{DBLP:journals/jmlr/RaffelSRLNMZLL20}
Colin Raffel, Noam Shazeer, Adam Roberts, Katherine Lee, Sharan Narang, Michael Matena, Yanqi Zhou, Wei Li, and Peter~J. Liu. 2020.
\newblock \href {http://jmlr.org/papers/v21/20-074.html} {Exploring the limits of transfer learning with a unified text-to-text transformer}.
\newblock \emph{J. Mach. Learn. Res.}, 21:140:1--140:67.

\bibitem[{Raheja et~al.(2023)Raheja, Kumar, Koo, and Kang}]{raheja2023coedit}
Vipul Raheja, Dhruv Kumar, Ryan Koo, and Dongyeop Kang. 2023.
\newblock \href {https://arxiv.org/abs/2305.09857} {Coedit: Text editing by task-specific instruction tuning}.

\bibitem[{Rajbhandari et~al.(2020)Rajbhandari, Rasley, Ruwase, and He}]{DBLP:conf/sc/RajbhandariRRH20}
Samyam Rajbhandari, Jeff Rasley, Olatunji Ruwase, and Yuxiong He. 2020.
\newblock \href {https://doi.org/10.1109/SC41405.2020.00024} {Zero: memory optimizations toward training trillion parameter models}.
\newblock In \emph{Proceedings of the International Conference for High Performance Computing, Networking, Storage and Analysis, {SC} 2020, Virtual Event / Atlanta, Georgia, USA, November 9-19, 2020}, page~20. {IEEE/ACM}.

\bibitem[{Ravanelli et~al.(2021)Ravanelli, Parcollet, Plantinga, Rouhe, Cornell, Lugosch, Subakan, Dawalatabad, Heba, Zhong et~al.}]{ravanelli2021speechbrain}
Mirco Ravanelli, Titouan Parcollet, Peter Plantinga, Aku Rouhe, Samuele Cornell, Loren Lugosch, Cem Subakan, Nauman Dawalatabad, Abdelwahab Heba, Jianyuan Zhong, et~al. 2021.
\newblock Speechbrain: A general-purpose speech toolkit.
\newblock \emph{arXiv preprint arXiv:2106.04624}.

\bibitem[{Renals et~al.(2007)Renals, Hain, and Bourlard}]{renals07_ami}
Steve Renals, Thomas Hain, and Herve Bourlard. 2007.
\newblock \href {https://doi.org/10.1109/ASRU.2007.4430116} {{Recognition and understanding of meetings the AMI and AMIDA projects}}.
\newblock In \emph{2007 IEEE Workshop on Automatic Speech Recognition and Understanding (ASRU)}, pages 238--247.

\bibitem[{Shah and {de Melo}(2020)}]{ShahDeMelo2020TypographicalErrorCorrection}
Kshitij Shah and Gerard {de Melo}. 2020.
\newblock Correcting the autocorrect: Context-aware typographical error correction via training data augmentation.
\newblock In \emph{Proceedings of the 12th {Language Resources and Evaluation Conference} ({LREC} 2020)}, Paris, France.

\bibitem[{Shazeer et~al.(2017)Shazeer, Mirhoseini, Maziarz, Davis, Le, Hinton, and Dean}]{DBLP:conf/iclr/ShazeerMMDLHD17}
Noam Shazeer, Azalia Mirhoseini, Krzysztof Maziarz, Andy Davis, Quoc~V. Le, Geoffrey~E. Hinton, and Jeff Dean. 2017.
\newblock \href {https://openreview.net/forum?id=B1ckMDqlg} {Outrageously large neural networks: The sparsely-gated mixture-of-experts layer}.
\newblock In \emph{5th International Conference on Learning Representations, {ICLR} 2017, Toulon, France, April 24-26, 2017, Conference Track Proceedings}. OpenReview.net.

\bibitem[{Srivastav et~al.(2023)Srivastav, Majumdar, Koluguri, Moumen, Gandhi, Team, Team, and Team}]{open-asr-leaderboard}
Vaibhav Srivastav, Somshubra Majumdar, Nithin Koluguri, Adel Moumen, Sanchit Gandhi, Hugging~Face Team, Nvidia~NeMo Team, and SpeechBrain Team. 2023.
\newblock \href {https://huggingface.co/spaces/huggingface.co/spaces/open-asr-leaderboard/leaderboard} {Open automatic speech recognition leaderboard}.
\newblock \emph{Hugging Face}.

\bibitem[{Sukhbaatar et~al.(2024)Sukhbaatar, Golovneva, Sharma, Xu, Lin, Rozi{\`{e}}re, Kahn, Li, Yih, Weston, and Li}]{DBLP:journals/corr/abs-2403-07816}
Sainbayar Sukhbaatar, Olga Golovneva, Vasu Sharma, Hu~Xu, Xi~Victoria Lin, Baptiste Rozi{\`{e}}re, Jacob Kahn, Daniel Li, Wen{-}tau Yih, Jason Weston, and Xian Li. 2024.
\newblock \href {https://doi.org/10.48550/ARXIV.2403.07816} {Branch-train-mix: Mixing expert llms into a mixture-of-experts {LLM}}.
\newblock \emph{CoRR}, abs/2403.07816.

\bibitem[{Team et~al.(2024)Team, Mesnard, Hardin, Dadashi, Bhupatiraju, Pathak, Sifre, Rivi{\`e}re, Kale, Love et~al.}]{team2024gemma}
Gemma Team, Thomas Mesnard, Cassidy Hardin, Robert Dadashi, Surya Bhupatiraju, Shreya Pathak, Laurent Sifre, Morgane Rivi{\`e}re, Mihir~Sanjay Kale, Juliette Love, et~al. 2024.
\newblock Gemma: Open models based on gemini research and technology.
\newblock \emph{arXiv preprint arXiv:2403.08295}.

\bibitem[{Valaki et~al.(2004)Valaki, Maestu, Simos, Zhang, Fernandez, Amo, Ortiz, and Papanicolaou}]{valaki2004cortical}
CE~Valaki, F~Maestu, PG~Simos, W~Zhang, A~Fernandez, CM~Amo, TM~Ortiz, and AC~Papanicolaou. 2004.
\newblock Cortical organization for receptive language functions in chinese, english, and spanish: a cross-linguistic meg study.
\newblock \emph{Neuropsychologia}, 42(7):967--979.

\bibitem[{Vaswani et~al.(2017)Vaswani, Shazeer, Parmar, Uszkoreit, Jones, Gomez, Kaiser, and Polosukhin}]{vaswani2017attention}
Ashish Vaswani, Noam Shazeer, Niki Parmar, Jakob Uszkoreit, Llion Jones, Aidan~N Gomez, {\L}ukasz Kaiser, and Illia Polosukhin. 2017.
\newblock Attention is all you need.
\newblock \emph{Advances in neural information processing systems}, 30.

\bibitem[{Vink et~al.(2020)Vink, Gladwin, Geeraerts, Pas, Bos, Hofstee, Durston, and Vollebergh}]{vink2020towards}
Matthijs Vink, Thomas~Edward Gladwin, Sanne Geeraerts, Pascal Pas, Dienke Bos, Marissa Hofstee, Sarah Durston, and Wilma Vollebergh. 2020.
\newblock Towards an integrated account of the development of self-regulation from a neurocognitive perspective: A framework for current and future longitudinal multi-modal investigations.
\newblock \emph{Developmental Cognitive Neuroscience}, 45:100829.

\bibitem[{Wang et~al.(2021)Wang, Riviere, Lee, Wu, Talnikar, Haziza, Williamson, Pino, and Dupoux}]{wang21_voxpopuli}
Changhan Wang, Morgane Riviere, Ann Lee, Anne Wu, Chaitanya Talnikar, Daniel Haziza, Mary Williamson, Juan Pino, and Emmanuel Dupoux. 2021.
\newblock \href {https://doi.org/10.18653/v1/2021.acl-long.80} {{VoxPopuli: A Large-Scale Multilingual Speech Corpus for Representation Learning, Semi-Supervised Learning and Interpretation}}.
\newblock In \emph{Proceedings of the 59th Annual Meeting of the Association for Computational Linguistics and the 11th International Joint Conference on Natural Language Processing (Volume 1: Long Papers)}, pages 993--1003, Online. Association for Computational Linguistics.

\bibitem[{Wang et~al.(2020)Wang, Wu, and Pino}]{DBLP:journals/corr/abs-2007-10310}
Changhan Wang, Anne Wu, and Juan~Miguel Pino. 2020.
\newblock \href {https://arxiv.org/abs/2007.10310} {Covost 2: {A} massively multilingual speech-to-text translation corpus}.
\newblock \emph{CoRR}, abs/2007.10310.

\bibitem[{Watanabe et~al.(2018)Watanabe, Hori, Karita, Hayashi, Nishitoba, Unno, Soplin, Heymann, Wiesner, Chen et~al.}]{watanabe2018espnet}
Shinji Watanabe, Takaaki Hori, Shigeki Karita, Tomoki Hayashi, Jiro Nishitoba, Yuya Unno, Nelson Enrique~Yalta Soplin, Jahn Heymann, Matthew Wiesner, Nanxin Chen, et~al. 2018.
\newblock Espnet: End-to-end speech processing toolkit.
\newblock \emph{arXiv preprint arXiv:1804.00015}.

\bibitem[{Watanabe et~al.(2017)Watanabe, Hori, Kim, Hershey, and Hayashi}]{watanabe2017hybrid}
Shinji Watanabe, Takaaki Hori, Suyoun Kim, John~R Hershey, and Tomoki Hayashi. 2017.
\newblock Hybrid ctc/attention architecture for end-to-end speech recognition.
\newblock \emph{IEEE Journal of Selected Topics in Signal Processing}, 11(8):1240--1253.

\bibitem[{Wolf et~al.(2020)Wolf, Debut, Sanh, Chaumond, Delangue, Moi, Cistac, Rault, Louf, Funtowicz, Davison, Shleifer, von Platen, Ma, Jernite, Plu, Xu, Scao, Gugger, Drame, Lhoest, and Rush}]{wolf-etal-2020-transformers}
Thomas Wolf, Lysandre Debut, Victor Sanh, Julien Chaumond, Clement Delangue, Anthony Moi, Pierric Cistac, Tim Rault, Rémi Louf, Morgan Funtowicz, Joe Davison, Sam Shleifer, Patrick von Platen, Clara Ma, Yacine Jernite, Julien Plu, Canwen Xu, Teven~Le Scao, Sylvain Gugger, Mariama Drame, Quentin Lhoest, and Alexander~M. Rush. 2020.
\newblock \href {https://www.aclweb.org/anthology/2020.emnlp-demos.6} {Transformers: State-of-the-art natural language processing}.
\newblock In \emph{Proceedings of the 2020 Conference on Empirical Methods in Natural Language Processing: System Demonstrations}, pages 38--45, Online. Association for Computational Linguistics.

\bibitem[{Yang et~al.(2023)Yang, Gu, Liu, Ghosh, Bulyko, and Stolcke}]{yang2023generative}
Chao-Han~Huck Yang, Yile Gu, Yi-Chieh Liu, Shalini Ghosh, Ivan Bulyko, and Andreas Stolcke. 2023.
\newblock Generative speech recognition error correction with large language models and task-activating prompting.
\newblock In \emph{2023 IEEE Automatic Speech Recognition and Understanding Workshop (ASRU)}, pages 1--8. IEEE.

\bibitem[{Yang et~al.(2021)Yang, Liu, Gandhe, Gu, Raju, Filimonov, and Bulyko}]{yang2021multi}
Chao-Han~Huck Yang, Linda Liu, Ankur Gandhe, Yile Gu, Anirudh Raju, Denis Filimonov, and Ivan Bulyko. 2021.
\newblock Multi-task language modeling for improving speech recognition of rare words.
\newblock In \emph{2021 IEEE Automatic Speech Recognition and Understanding Workshop (ASRU)}, pages 1087--1093. IEEE.

\bibitem[{Yasunaga et~al.(2021)Yasunaga, Leskovec, and Liang}]{yasunaga2021lm}
Michihiro Yasunaga, Jure Leskovec, and Percy Liang. 2021.
\newblock Lm-critic: Language models for unsupervised grammatical error correction.
\newblock \emph{arXiv preprint arXiv:2109.06822}.

\bibitem[{Zatorre and Gandour(2008)}]{zatorre2008neural}
Robert~J Zatorre and Jackson~T Gandour. 2008.
\newblock Neural specializations for speech and pitch: moving beyond the dichotomies.
\newblock \emph{Philosophical Transactions of the Royal Society B: Biological Sciences}, 363(1493):1087--1104.

\end{thebibliography}








\clearpage
\appendix
\section{Appendix}

\paragraph{Training Details}
We fine-tune the model for 3 epochs using the AdamW optimizer \citep{DBLP:conf/iclr/LoshchilovH19} with a learning rate of $1e-4$ and a weight decay of $0.01$. We use a cosine learning rate scheduler with a warmup ratio of $0.1$ and a gradient clipping threshold of $1.0$.
For the expert-dataset mapping, we randomly assign each dataset to one of the 8 experts in the Mixtral model. This random assignment serves as a strong baseline and allows us to focus on the effectiveness of the task-guided expert assignment approach. We leave the exploration of more advanced expert assignment strategies for future work.
To efficiently train the large-scale model, we leverage DeepSpeed Zero \citep{DBLP:conf/sc/RajbhandariRRH20} for memory optimization and Hugging Face Transformers \citep{wolf-etal-2020-transformers} for model implementation. 

\textit{Translation Tasks.} As an extra zero-shot textual correction setup, we evaluate \methodwithspace on machine translation (MT) of WMT'20 for Japanese and Chinese \citep{barrault-etal-2020-findings}.
We use the BLEU score \citep{DBLP:conf/acl/PapineniRWZ02} as the evaluation metric for ST (with training and test) and MT (zero-shot). 

\textit{Grammar Correction Tasks.} These text error correction (TEC) tasks focus on correcting grammatical errors and improving the overall coherence of the text, making them suitable for evaluating the effectiveness of our model in handling TEC-related editing instructions. We use the word error rate as the evaluation metric.

\textit{OCR Tasks.} The dataset includes original texts with varying numbers of OCR mistakes and their corresponding corrected versions. To evaluate our model, we take the first 1,000 characters of both the input text with OCR errors and the ground-truth corrected text. We use the WER as the evaluation metric.

\paragraph{Mixture of Experts Background}
Mixture-of-experts (MoE) \citep{DBLP:conf/iclr/ShazeerMMDLHD17} is a machine learning concept that employs multiple expert layers, each of which specializes in solving a specific subtask. The experts then work together to solve the entire task at hand. Recently, MoE has been widely applied to large-scale distributed Deep Learning models by using a cross-GPU layer that exchanges hidden features from different GPUs \citep{DBLP:conf/iclr/LepikhinLXCFHKS21,DBLP:journals/jmlr/FedusZS22}.
The MoE approach is differentiated from existing scale-up approaches for DNNs, such as increasing the depth or width of DNNs, in terms of its high cost-efficiency. Specifically, adding more model parameters (experts) in MoE layers does \textbf{not} increase the computational cost per token at inference time. Thus, MoE has been studied for scaling the models to trillion-size parameters in NLP \citep{DBLP:journals/jmlr/FedusZS22}.

\begin{figure}[h!]
\centering
\includegraphics[width=0.48\textwidth]{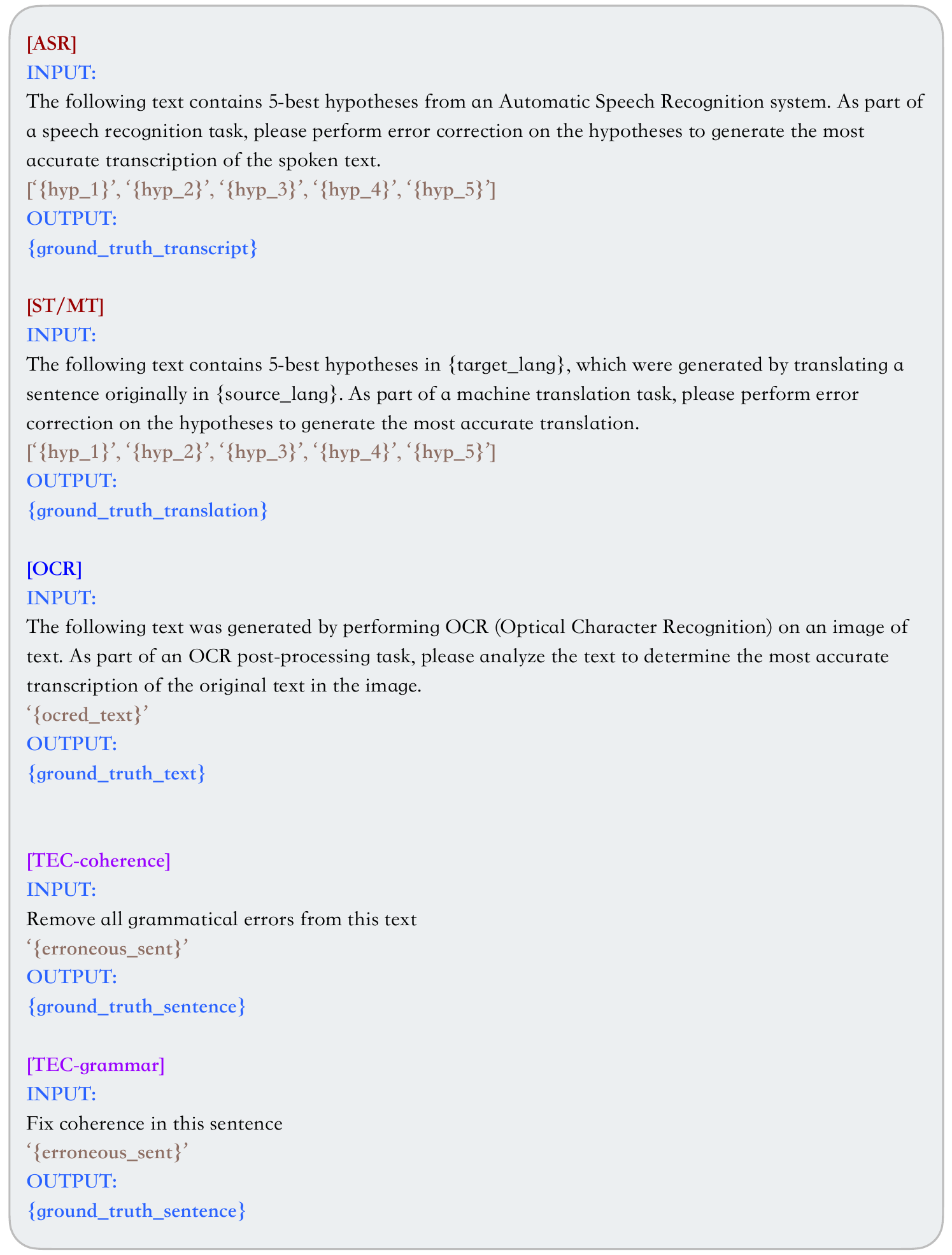}
\caption{Example prompts of various correction tasks using Automatic Speech Recognition (ASR), Machine Translation (MT), Speech Translation (ST), Optical Character Recognition (OCR), and Textual Error Correction (TEC).}
\label{fig:task_prompt}
\end{figure}

\paragraph{Prompt Format}
We provide detailed correction example per \textsc{[TASK]} and actual prompt format of \textsc{INPUT:} used in the our experiments for qualitative studies as shown in Figure~\ref{fig:task_prompt}. For instance, each task will have a specific task-activation prompt format, where ASR, ST, and MT would be based on the sampling or beam search results. On the other hand, OCR and TEC will use input texts for end-to-end mapping.


\paragraph{Correction Examples}
\label{apx:sec:example}

\begin{figure}[h!]
\centering
\includegraphics[width=0.48\textwidth]{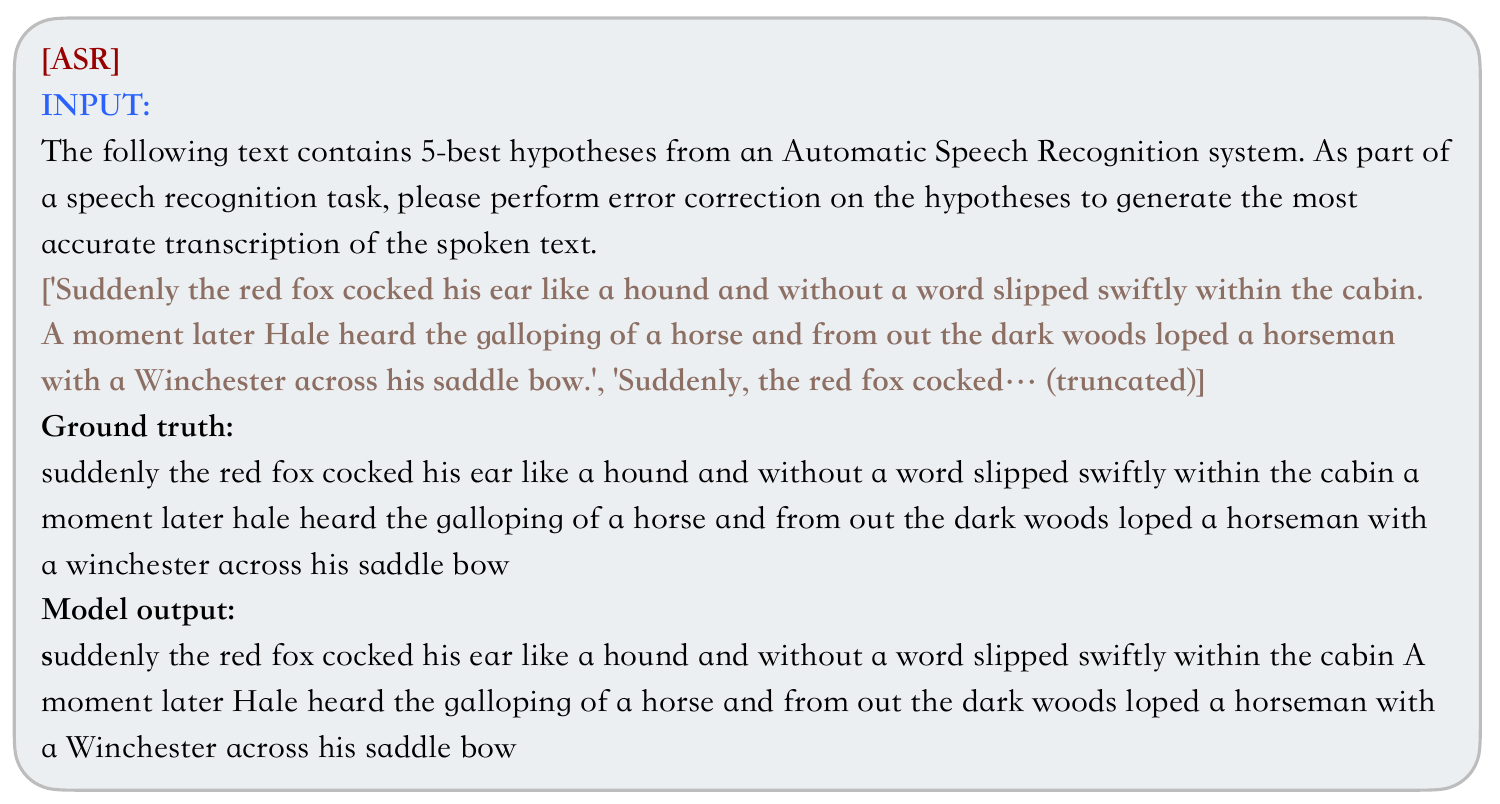}
\caption{Examples of \methodwithspace outputs for asr error correction task in SPGISpeech~\cite{oneill21_kensho}.}
\label{fig:example:asr}
\end{figure}

\begin{figure}[h!]
\centering
\includegraphics[width=0.48\textwidth]{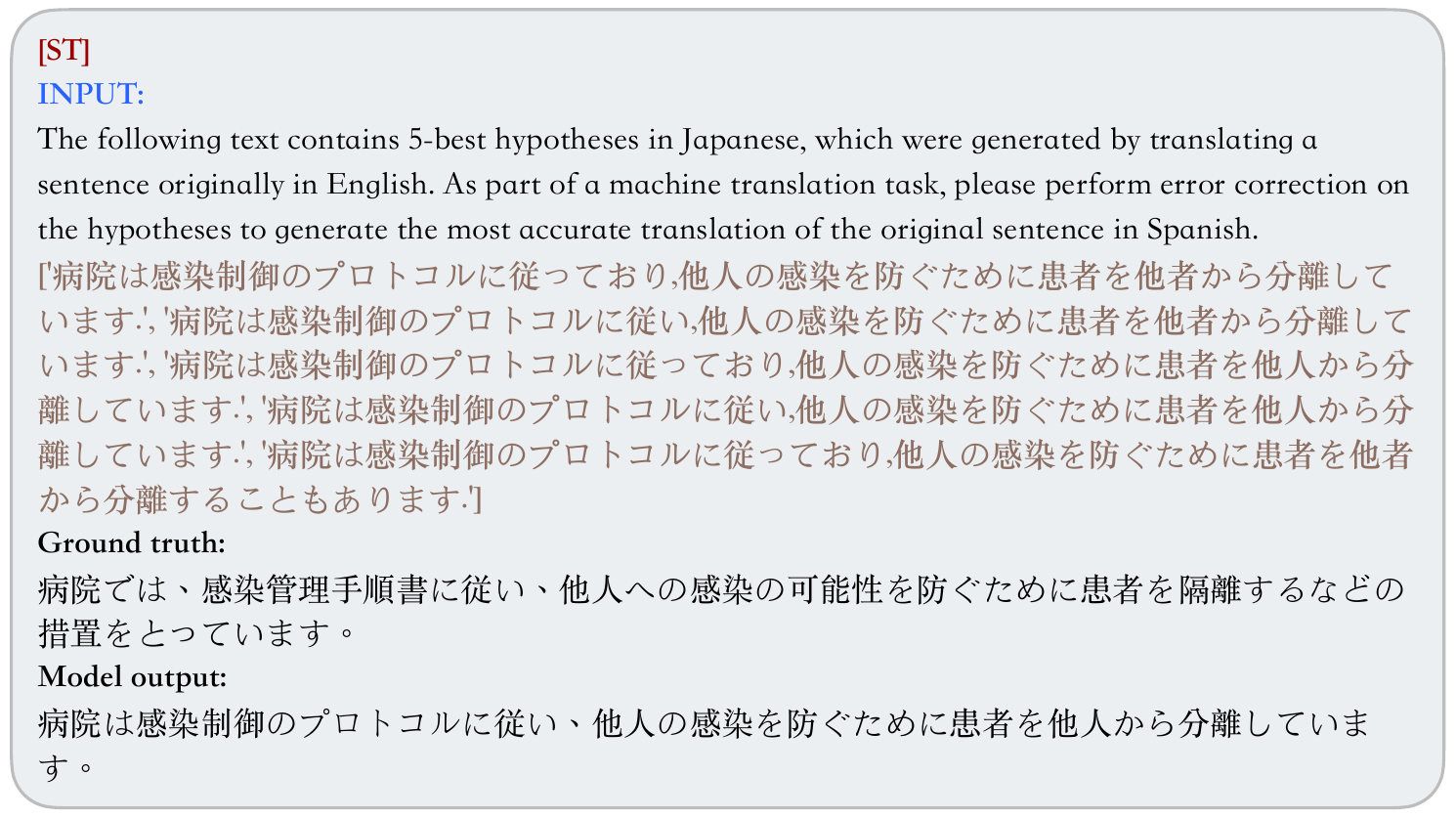}
\caption{Examples of \methodwithspace outputs for speech translation correction task in FLEURS \cite{DBLP:conf/slt/ConneauMKZADRRB22}.}
\label{fig:example:st}
\end{figure}

\begin{figure}[h!]
\centering
\includegraphics[width=0.48\textwidth]{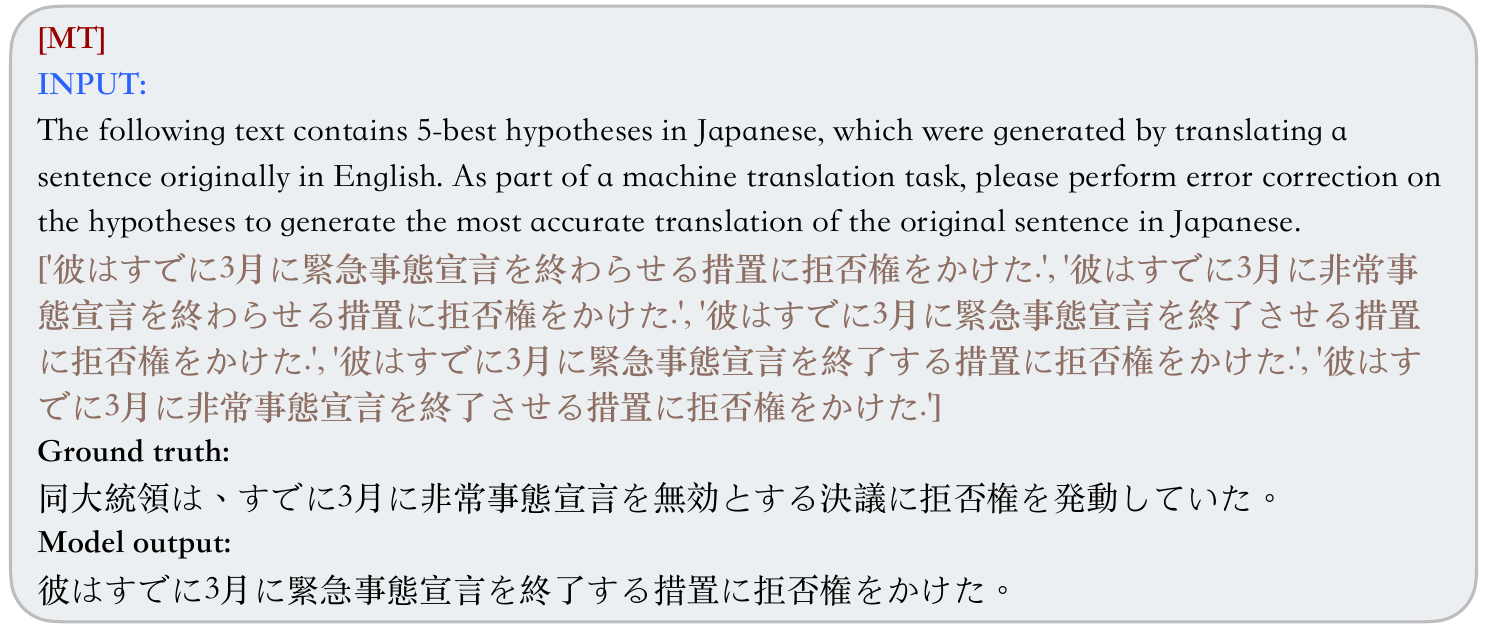}
\caption{Examples of \methodwithspace outputs for machine translation correction task in WMT20 \cite{barrault-etal-2020-findings}.}
\label{fig:example:mt}
\end{figure}

\begin{figure}[ht!]
\centering
\includegraphics[width=0.48\textwidth]{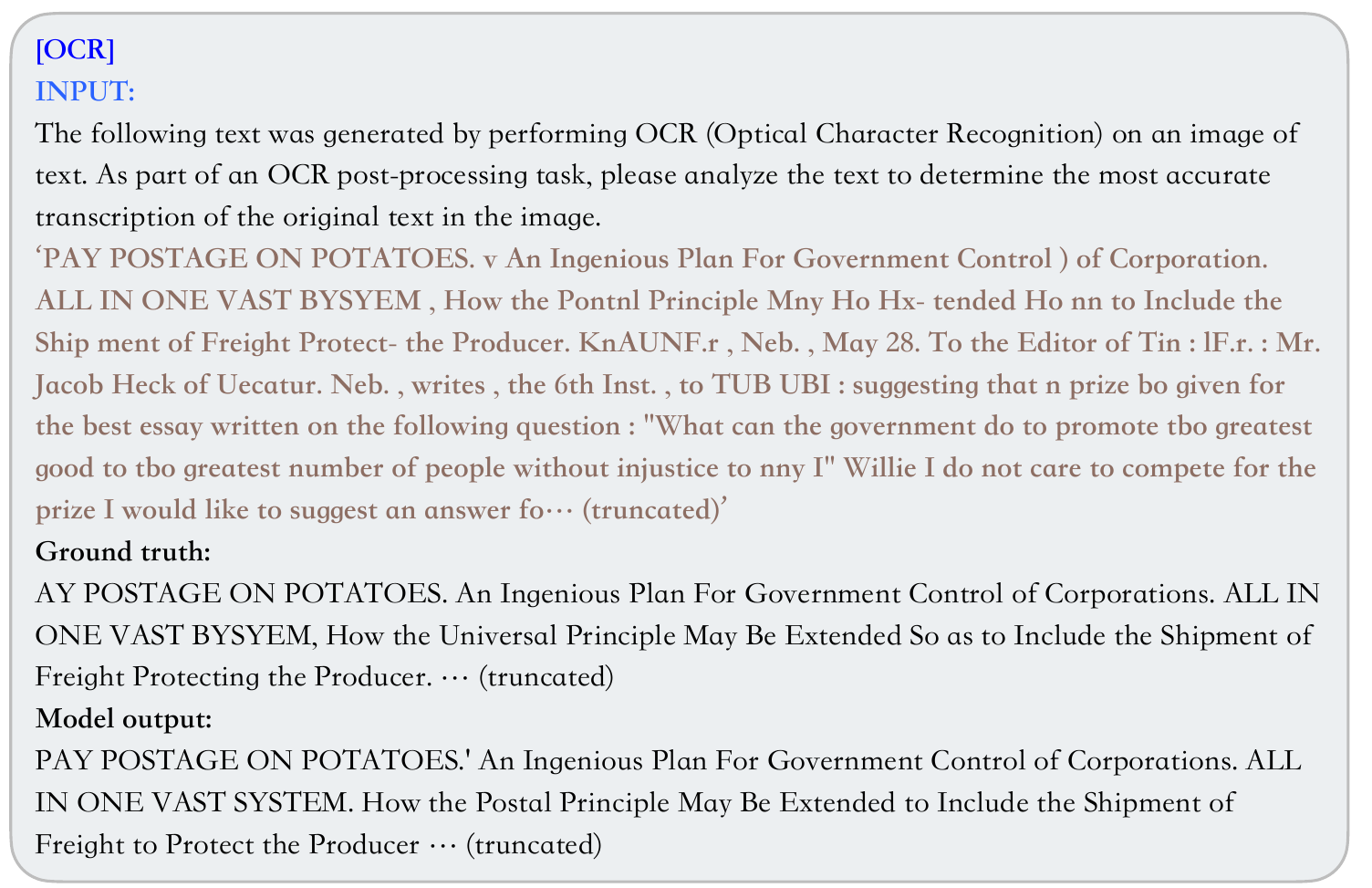}
\caption{Examples of \methodwithspace outputs for OCR correction task in \href{https://huggingface.co/datasets/PleIAs/Post-OCR-Correction}{PleIAs/Post-OCR-Correction}.}
\label{fig:example:ocr}
\end{figure}

\begin{figure}[h!]
\centering
\includegraphics[width=0.48\textwidth]{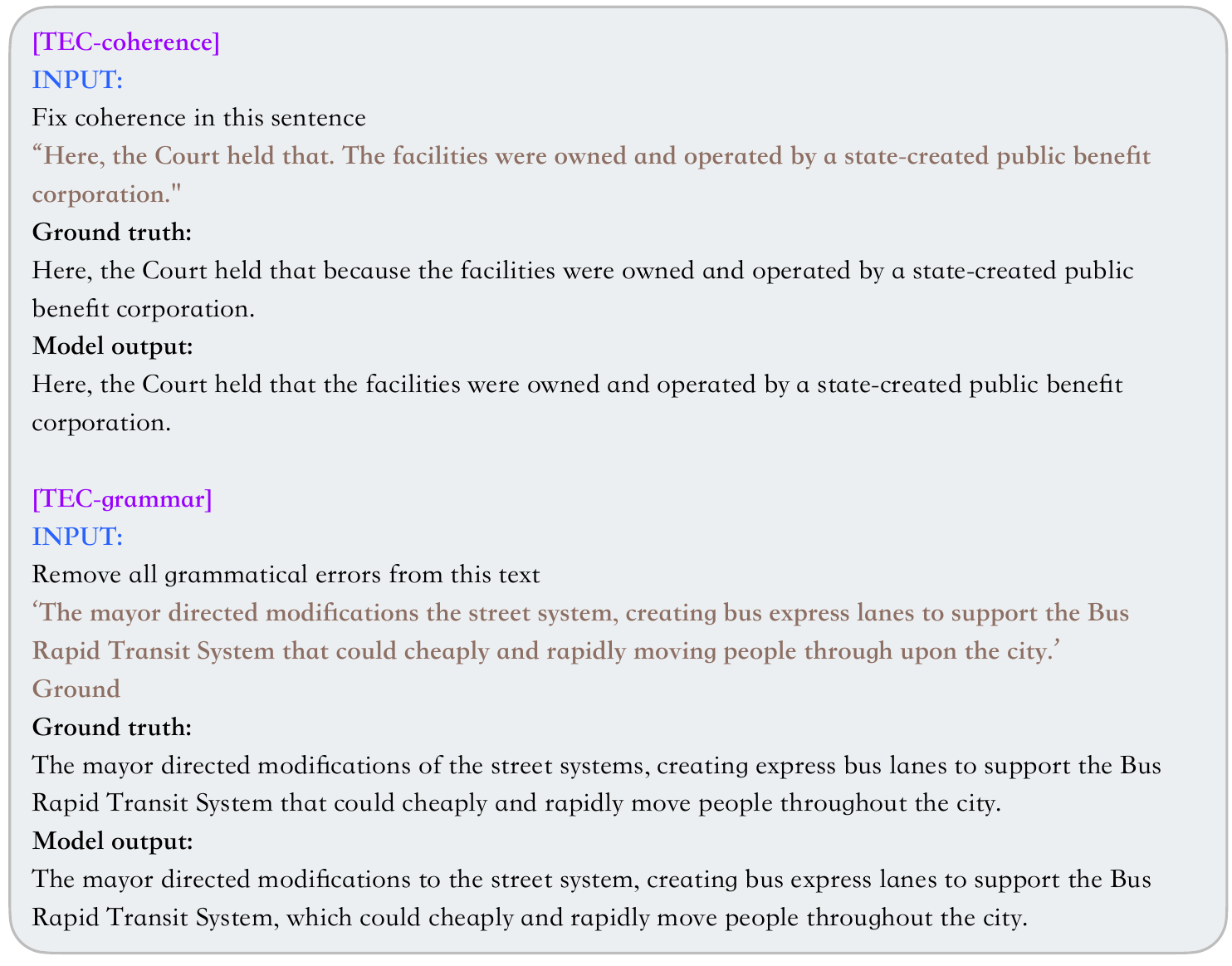}
\caption{Examples of \methodwithspace outputs for textual error correction (TEC) tasks in CoEdIT \cite{raheja2023coedit}.}
\label{fig:example:gec}
\end{figure}

We randomly select post-recognition example by \methodwithspace. In Figure~\ref{fig:example:asr}, a long form ASR output has been selected and it remain the top 1-best correction with~\method. or the ST and MT correction result in Figure~\ref{fig:example:st} and in Figure~\ref{fig:example:mt}, although the post-\methodwithspace corrected output does not perfectly align with the ground truth, it boosts the general semantic meaning, as reviewed by native speakers. Meanwhile, the OCR and TEC correction results in Figures ~\ref{fig:example:ocr} and ~\ref{fig:example:gec} demonstrate various types of corrections, such as pattern-wise character misrecognition and understanding-based coherence improvements.

\begin{figure}[h!]
\centering
\includegraphics[width=0.48\textwidth]{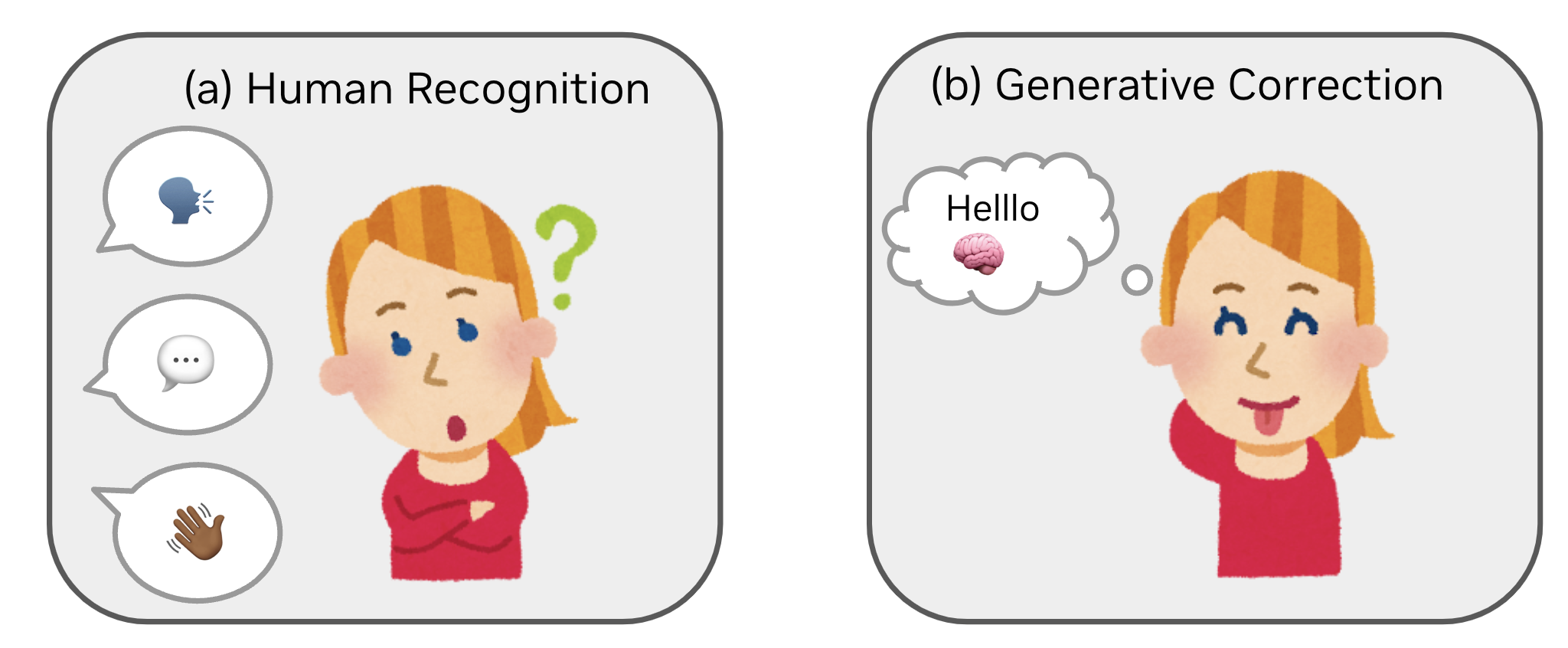}
\caption{Examples of (a) Human recognition given different input modalities, including audio, text, and visual patterns; (b) generative inference and correction~\cite{marshall2015deaf, levinson2010time} to understand the recognition results.}
\label{fig:human:cog}
\end{figure}

\paragraph{Additional Discussion on Human Recognition from Speech and Text Inputs}
Human recognition (e.g., speech, optical character, text translation) and has naturally evolved to excel at recognizing and understanding speech in a wide range of real-world scenarios \cite{he2019streaming, deng2013recent}. However, the field of automatic speech recognition (ASR) has traditionally concentrated on training and evaluating models on specific datasets \cite{ chan2016listen, watanabe2017hybrid}. These models have shown limited adaptability to new environments \cite{yang2021multi,du2016ustc, hu2024large}, leading to decreased accuracy and practicality in real-world settings.
Recognizing the challenges posed by single dataset models and the availability of diverse datasets collected over time, unified models are being developed that merge information from multiple datasets into a single framework \cite{barrault2023seamlessm4t}. While Grammatical Error Correction (TEC) has been actively explored \cite{yang2023generative}, ASR error correction is distinct due to the arbitrariness of spoken language \cite{aksenova2021might}, requiring efforts from both speech, NLP, and cognitive science communities as one human recognition example shown in Figure~\ref{fig:human:cog}.

\paragraph{task-guided Inference for Mixture of Expert Models} 
During inference, the Neko-model utilizes top-2 expert routing, instead of just top-1. Our pilot studies showed that top-1 routing indeed led to worse performance due to limited knowledge sharing.

Using more than two experts (e.g., top-3 or higher) diverged from the training setup and increased inference costs (ranging from $23.5$\% to $75.5$\%) without significant gain (i.e., a relative difference of less than $0.06$\%).

\paragraph{Future Model Maintenance Plan and ASR Community}

\begin{figure}[h!]
\centering
\includegraphics[width=0.48\textwidth]{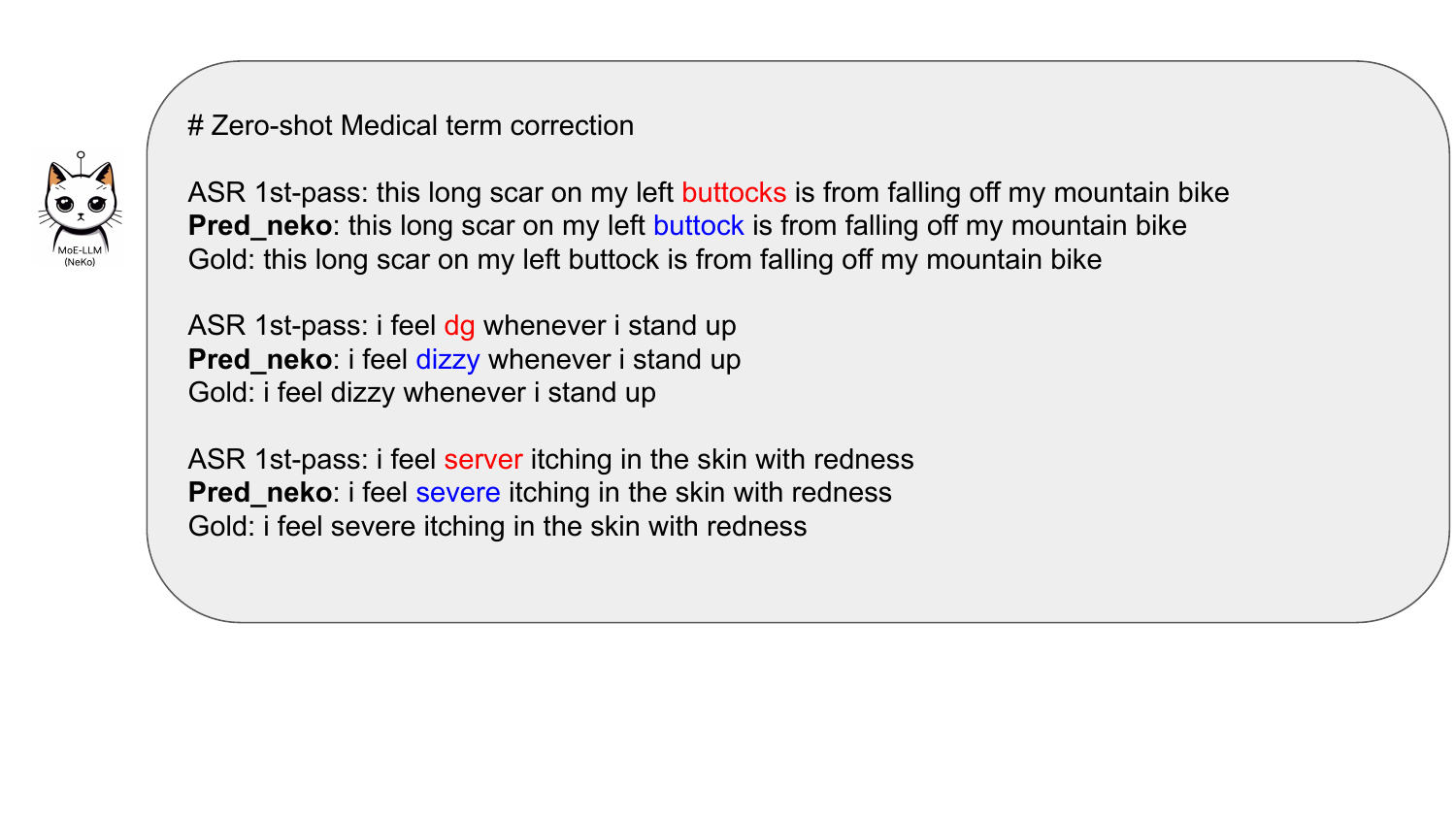}
\includegraphics[width=0.48\textwidth]{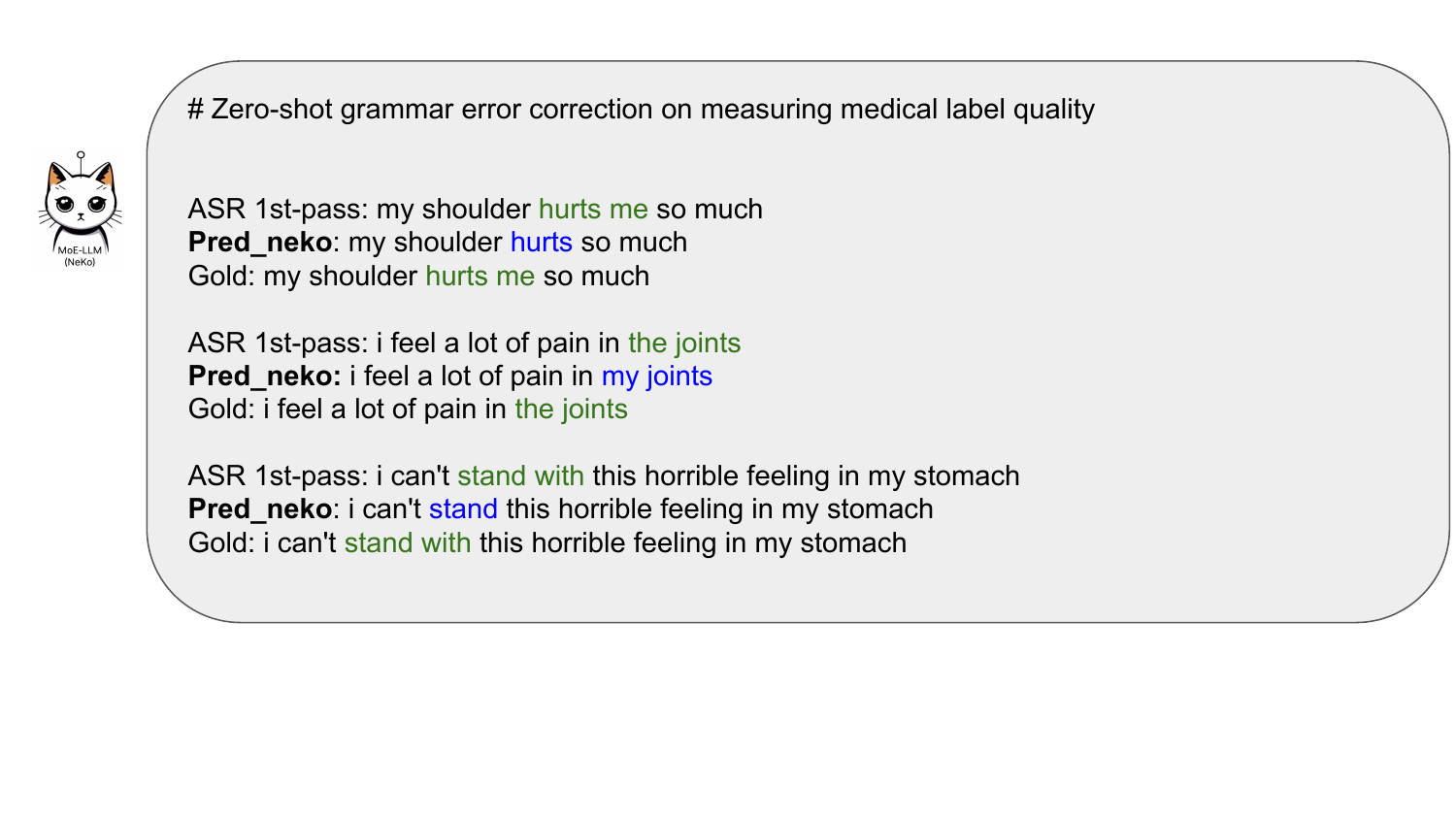}
\caption{We provide medical post-ASR recognition correction on the Medical-ASR-EN dataset (\url{https://huggingface.co/datasets/jarvisx17/Medical-ASR-EN}), where NeKo demonstrates the ability to (1) refine clinically related term errors and (2) correct grammar format.}
\label{fig:med:1}
\end{figure}

For ASR tasks, we used Canary-v0, Whisper-seires, and SeamlessM4T to decode textual hypotheses data. For Whisper, we included it as a widely-used baseline, but our key comparisons are to other GEC methods also using Whisper (e.g. GenTranslate). Open eco-system, including ESPnet~\citep{watanabe2018espnet} and SpeechBrain~\citep{ravanelli2021speechbrain} models, are also our interests to be adapted as first-pass ASR in the open code base. This will provide a more comprehensive evaluation across model types. In general, NeKo's post-ASR correction improvements are consistent across datasets and first-pass models, suggesting the benefits generalize beyond model-specific (\textit{i.e.}, Canary, Whisper, or SeamlessM4T) 's strengths as the initial medical term correction results shown in Figure~\ref{fig:med:1}.

\paragraph{Emergent Unseen Task Zero-Shot Performance} We investigate \method's generalization capabilities to unseen tasks using an additional synthetic typographical error correction dataset~\citep{ShahDeMelo2020TypographicalErrorCorrection}. This dataset is derived from the IMDb test split, featured low noise levels (3.75\% character error rate) with corruption applied using algorithms proposed in~\citep{ShahDeMelo2020TypographicalErrorCorrection}. Our evaluation focused on zero-shot and five-shot learning scenarios to assess the adaptability of various models without and with minimal task-specific training. In the zero-shot scenario, where models were prompted to switch from an ASR task to typo correction without additional training, the challenge proved significant. The models, including the advanced Claude-Opus, yielded WERs above 30\%. The predictions were markedly irrelevant to the ground truth, highlighting the difficulty of adapting to typo correction without specific fine-tuning. This finding prompts further investigation into efficient and effective training techniques for generalizing model capabilities across diverse linguistic tasks. In the five-shot scenario, all models improved against the corrupted baseline with Claude-Opus performing best. Notably, \method \space outperformed GPT-3.5-Turbo, indicating some affinity towards this task. 

\paragraph{Task-Specific Fine-Tuning}
The \textsc{NeKo} model employs task-guided MoE fine-tuning, where each expert is assigned to a specific dataset. This approach may lead to overfitting to the specific characteristics of the training datasets even though knowledge could be shared. As a result, the model's performance might degrade when applied to new tasks or datasets that were not part of the training set, limiting its adaptability. Investigating more dynamic and adaptive fine-tuning strategies that can generalize better across unseen tasks and datasets would be beneficial. Techniques such as meta-learning or continual learning  could be explored to enhance the model's adaptability and robustness.

\paragraph{Future Connections to In-Context and Auto-Agent Learning with \method}
Integrating in-context learning (ICL) with \textsc{NeKo} could enable the model to adapt to various error correction tasks by conditioning on input examples without requiring explicit fine-tuning. This approach is particularly beneficial in scenarios where obtaining large labeled datasets for fine-tuning is impractical. By leveraging ICL, \textsc{NeKo} could adapt to diverse error types and use in-context examples to correct errors specific to new domains or applications, thereby improving its generalizability to real-world data. Furthermore, ICL would allow the model to dynamically adjust its error correction strategies based on the input context, enhancing its robustness to varying error distributions.


\begin{table}[ht]
    \centering
    \caption{WER comparison of \methodwithspace against GPT-3.5-Turbo,  and Claude-Opus on the 5-shot IMDb typographical error correction dataset \citep{ShahDeMelo2020TypographicalErrorCorrection}. The baseline represents the WER between the corrupted text and the ground truth. Lower WER indicates better performance in correcting typographical errors.}
    \begin{tabular}{lcc}
        \toprule
        Model & WER \\
        \midrule
        Baseline (Corrupt vs Ground Truth) & 18.35\% \\
        GPT-3.5-Turbo (5-shots) & 12.72\% \\
        Claude-3-Sonnet  (5-shots) & 12.18\% \\
        Claude-3.5 Sonnet (5-shots) & 8.18\% \\
        \method-MoE (5-shots) & 11.62\% \\
        \bottomrule
    \end{tabular}
    \label{tab:wer_comparison}
\end{table}
\vspace{0.5cm}

\paragraph{Acknowledgment} 

The authors would like to acknowledge the Taipei-1 high-performance computing support provided by the NVIDIA Taiwan Research and Development Center (TRDC) project, and the initial feedback from NVIDIA’s Eric Kang, Taejin Park, Boyi Li, and Yejin Choi.

\clearpage


\end{document}